\documentclass[11pt]{report}
\usepackage[bindingoffset=0cm,left=4cm,right=4cm,top=3.5cm,bottom=3.5cm]{geometry}
\pdfoutput=1

\usepackage{amsmath}
\usepackage{amsfonts}
\usepackage{amssymb}
\usepackage{algorithm}
\usepackage{algorithmic}
\usepackage{url}
\usepackage{color}
\usepackage{graphicx}
\usepackage{listings}
\usepackage[Lenny]{fncychap}
\usepackage{fancyhdr}
\usepackage[pdfstartview = FitH]{hyperref}

\hypersetup{
pdfborder = 0 0 0,
pdftitle = {Vincent Etter - Master Thesis - Semantic Vector Machines},
pdfkeywords = {vincent, etter, machine learning, machine translation, embedding, semantic compression, semantic extraction, semantic, vector, machines, epfl, nec labs, princeton},
pdfauthor = {Vincent Etter},
pdfcreator = {\LaTeX\ with package hyperref},
}

\definecolor{gray}{gray}{0.95}

\author{Vincent Etter}
\title{Semantic Vector Machines}
\begin{document}

\input{lua.def}
\lstset{ %
language=lua,                
basicstyle=\footnotesize,       
numbers=none,                   
numberstyle=\footnotesize,      
stepnumber=1,                   
numbersep=5pt,                  
backgroundcolor=\color{gray},  
showspaces=false,               
showstringspaces=false,         
showtabs=false,                 
frame=single,                   
tabsize=2,              
captionpos=b,                   
breaklines=true,        
breakatwhitespace=false,    
escapeinside={\%}{)}          
}

\setlength{\headheight}{15pt}
 
\renewcommand{\chaptermark}[1]{\markboth{#1}{}}

\fancyhf{}
\fancyhead[LE,RO,RE,LO]{\thepage}
\fancyhead[LE]{\textit{\nouppercase{\rightmark}}}
\fancyhead[RE]{\textit{\nouppercase{\leftmark}}}
\fancyhead[LO]{\textit{\nouppercase{\leftmark}}}
\fancyhead[RO]{\textit{\nouppercase{\rightmark}}}
\cfoot{\thepage}

\fancypagestyle{plain}{ %
\fancyhf{} 
\renewcommand{\headrulewidth}{0pt} 
\renewcommand{\footrulewidth}{0pt}
\cfoot{\thepage}}

\pagestyle{empty}

\titlepage

\begin{center}
Winter 2008-2009

\vspace{4cm}

\begin{tabular}{c}
\hline \\
\LARGE {\textsc{Semantic Vector}} \\
\LARGE {\textsc{Machines}} \\
----- \\
\large{\textsc{Master Thesis}} \\
\large{\textsc{in Communication Systems}} \vspace{3mm} \\
\hline 
\end{tabular}

\vspace{3.5cm}
\large{\textbf{Vincent Etter}} \\
\vspace{2mm}
\begin{normalsize}
R\'{e}s. Les Roseyres D12\\
1882 Gryon \\
Switzerland \vspace{1.5mm} \\
vincent.etter@gmail.com
\end{normalsize}

\vspace{3.5cm}

\begin{small}
\begin{tabular}{p{7cm}p{7cm}}
\footnotesize{Responsible Professor:} & \footnotesize{Supervisor:} \\
\textbf{Martin Hasler} & \textbf{Ronan Collobert} \\
Laboratory of Nonlinear Systems & Department of Machine Learning \\
\'{E}cole Polytechnique F\'{e}d\'{e}rale de Lausanne & NEC Laboratories America, Inc. \\
1015 Lausanne, VD & Princeton, NJ 08540 \\
Switzerland & USA \\
\end{tabular}
\end{small}

\end{center}
\cleardoublepage
\begin{abstract}
We first present our work in machine translation, during which we used aligned sentences to train a neural network to embed n-grams of different languages into an $d$-dimensional space, such that n-grams that are the translation of each other are close with respect to some metric. Good n-grams to n-grams translation results were achieved, but full sentences translation is still problematic. We realized that learning semantics of sentences and documents was the key for solving a lot of natural language processing problems, and thus moved to the second part of our work: sentence compression. We introduce a flexible neural network architecture for learning embeddings of words and sentences that extract their semantics, propose an efficient implementation in the \texttt{Torch} framework and present embedding results comparable to the ones obtained with classical neural language models, while being more powerful.
\end{abstract}
\cleardoublepage
\renewcommand{\abstractname}{Acknowledgements}

\begin{abstract}
First of all, I would like to thank NEC Labs for allowing me to come for six months to the United States and work on such an interesting project and in such an great environment. This was a wonderful experience, both on personal and professional levels. I met very interesting people at the lab, and had the opportunity to attend fascinating talks all along my project.

I would also like to thank:

Ronan Collobert, for his support in both my professional and personal life. He gave me great advice and feedback all along my project, and was always here when I needed support with the fantastic Torch framework he created. He also showed me some beautiful places in New Jersey and New York state, and even shared his love of Quebec during a great trip there.

Jason Weston, for his help during the first part of my project.

Prof. Martin Hasler, for accepting to supervise my Master Thesis from EPFL, taking the time to read my monthly reports and providing me with some helpful feedback.

\end{abstract}
\cleardoublepage

\pagestyle{fancy}

\makeatletter
\def\cleardoublepage{\clearpage\if@twoside \ifodd\c@page\else
    \hbox{}
    \thispagestyle{plain}
    \newpage
    \if@twocolumn\hbox{}\newpage\fi\fi\fi}
\makeatother \clearpage{\pagestyle{plain}\cleardoublepage}

\tableofcontents
\cleardoublepage

\chapter{Introduction}

This report covers our Master Thesis project at NEC Laboratories America, in Princeton, New Jersey, which was done from September 8th, 2008 to March 13th, 2009. This project was about Machine Learning and Natural Language Processing. Machine Learning is the field that studies the design and implementation of algorithms that allow computers to improve their performances over time based on data, whereas Natural Language Processing is a field of Computer Sciences concerned with the interaction between computers and human (natural) languages.
\\

At the intersection of these fields, many tasks are found: semantic role labeling, name entity recognition, etc. We focused on two of them, that are Machine Translation and Sentence Compression. Machine translation has become more and more popular in the last years, mostly because of the huge quantity of data available on the Internet in other languages. People want to be able to read the latest news on a Japanese website, or to understand a new recipe from the blog of a Peruvian chef. To do so, they need tools that will translate for them all these texts from various foreign languages to their mother tongue. We will try to address this problem in the first part of our project, by building an automated tool that learns how to translate from one language to another.
\\

Sentence compression, or more generally semantic extraction, is a field that has become even more needed than translation. Indeed, we are in an information era, where electronic devices are more than ever part of our lives, and generate huge amounts of data: positions from GPS receivers, images from cellphones and cameras, texts from blogs, temperatures from sensors, etc. With this continuous flow of data, it has become impossible for humans to interpret it, or even sort it. Thus, we need tools that are able to make sense out of this pile of unlabeled data, to organize it. Semantic extraction aims to extract a representation of some data (in our case, text data) that gives information about its meaning. We will try, in the second part of our project, to build such tool, that embeds any text into a $d$-dimensional space in a way that this representation contains all the features needed to get its meaning. With a tool like that, we would be able to easily categorize, sort and cluster tons of information in an unsupervised way.
\\

This report is organized as follows:

In chapter \ref{chap:theory}, we present some theoretical reminders on neural networks and embedding, for readers that are not familiar with these concepts.

In chapter \ref{chap:torch}, we introduce the programming framework we used to implement all the algorithms and neural networks we designed during this project: \texttt{Torch}.

In chapter \ref{chap:translation}, we describe the first part of our project, in which we build a neural network that learns to translate from a language to another, by embedding words and sentences into a $d$-dimensional space, using a corpus of aligned sentences in both languages. The obtained results are then presented, and we conclude this chapter by some considerations about the limitations of our model, and reasons for redefining the problem and trying a new approach.

In chapter \ref{chap:compression}, we discuss the second part of our project, that focused on sentence compression. We propose a new approach, that still relies on embedding, and present the two main problems associated with it. A solution to each of them is then presented, and the performances are compared to previous work in the field on language models.

Finally, in chapter \ref{chap:conclusion}, we conclude this project and share some propositions for future work.
\cleardoublepage
\chapter{Theory}
\label{chap:theory}

In this chapter, we  introduce some important concepts that were used during our project. They may not be directly related to each other, but will be referred to in the next chapters and should thus be familiar to the reader.

\section{Neural networks}
\label{sec:theory-nn}

A neural network is a mathematical model, based on biological observations, that can be used for several different tasks, the most common ones being classification and regression. It is built using layers of interconnected artificial neurons. These artificial neurons simply take some input values and return a weighted sum of them, sometimes after having applied a non-linear function to it. While a neural network with only one layer is quite simple, adding a second layer allows the network to approximate any function, assuming there are enough neurons per layer. Thus, neural networks can be seen as universal approximators.
\\

\begin{figure}
\begin{center}
\includegraphics[width=0.4\textwidth]{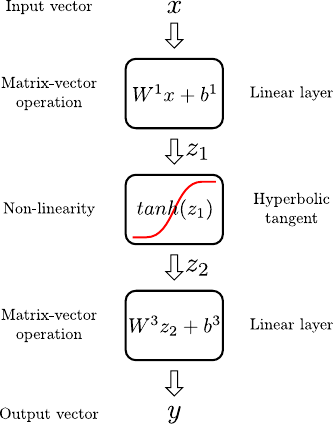}
\end{center}
\caption{Decomposition of a neural network into a sequence of matrix-vector operations and function applications.}
\label{fig:theory-nn}
\end{figure}

\subsection{A sequence of operations}

In general, we can express any network by a sequence of matrix-vector operations and applications of non-linear functions, as illustrated in figure \ref{fig:theory-nn}. For example, the following equation describes a network with $N$ inputs, $H$ neurons in the second layer (also called hidden layer) with a hyperbolic tangent transfer function, and $O$ outputs:

$$y = f_\theta(x) = W^3tanh(W^1x + b^1) + b^3$$

where $\theta = (W^1, b^1, W^3, b^3)$, $x \in \mathbb{R}^N$ the input of the network, $W^1 \in \mathbb{R}^H \times \mathbb{R}^N$ and $b^1 \in \mathbb{R}^H$ the weights and bias of the hidden layer, $W^3 \in \mathbb{R}^O \times \mathbb{R}^H$ and $b^3 \in \mathbb{R}^O$ the weights and bias of the output layer.

\subsection{Backpropagation}
\label{subsec:backprop}

When training a neural network, we want to optimize it with respect to a function $R$, called the \textit{expected risk}:

$$R = \mathbb{E}_x[C_w(x)]$$

where $C_w(x)$ is a \textit{loss function} that measures the performance of the network with respect to the given input $x$ and its current weights $w$. We cannot minimize the expected risk directly, as we do not know the distribution of $x$. However, it is proven that it is sufficient to minimize an approximation of the expected risk over a finite set of $N$ training samples:

$$R^* = \frac{1}{N} \sum_{i=1}^N C_w(x_i)$$

To do so, we usually use a gradient descent technique: for each sample, we estimate the loss and compute the gradient $\frac{\partial C}{\partial w^l_{jk}}$, that gives the error for each weight $jk$ of each layer $l$. Then, we simply apply a scaling factor $\gamma$ to this gradient, called learning rate, and subtract it to each weight. This gradient descent algorithm is a well-known optimization technique, which allows to find a local minima pretty quickly (depending on the learning rate).
\\

This means that we want to calculate $\frac{\partial C}{\partial w^l_{jk}}$, the gradient of each weight of a given layer, for all layers. This gradient can be rewritten as:

\begin{equation}
\frac{\partial C}{\partial w^l_{jk}} = \frac{\partial C}{\partial m^l} \frac{\partial m^l}{\partial w^l_{jk}}
\label{eq:1}
\end{equation}

where $m^l$ is the function applied at layer $l$ (e.g. a matrix-vector operation). The second term of the above equation is known, but we cannot compute the first one from scratch at each layer. However, we can rewrite it as:

\begin{equation}
\frac{\partial C}{\partial m^l} = \frac{\partial C}{\partial m^{l + 1}} \frac{\partial m^{l + 1}}{\partial m^l}
\label{eq:2}
\end{equation}

\begin{figure}
\begin{center}
\includegraphics[width=0.2\textwidth]{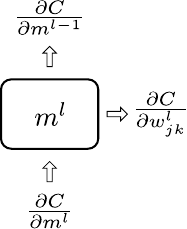}
\end{center}
\caption{Illustration of the backpropagation: each layer $l$ receives the gradient $\frac{\partial C}{\partial m^l}$ from the next layer, uses it to compute all $\frac{\partial C}{\partial w^l_{jk}}$ and backwards $\frac{\partial C}{\partial m^{l - 1}}$ to the previous layer to continue the chain.}
\label{fig:theory-backprop}
\end{figure}

where $m^{l + 1}$ is the operation applied at layer $l + 1$. As the first term is known for the last layer of the network (it is the gradient of the loss function), we can use this property to iteratively compute $\frac{\partial C}{\partial w^l_{jk}}$ at each layer $l$ and then propagate $\frac{\partial C}{\partial m^l}$ to the previous layer, to allow it to do its own computations. The procedure for each layer $l$, illustrated in figure \ref{fig:theory-backprop}, is thus:

\begin{enumerate}
\item Receive $\frac{\partial C}{\partial m^l}$ from next layer
\item Compute all $\frac{\partial C}{\partial w^l_{jk}}$ using equation \ref{eq:1}
\item Compute $\frac{\partial C}{\partial m^{l - 1}}$ using equation \ref{eq:2}
\item Propagate $\frac{\partial C}{\partial m^{l - 1}}$ to previous layer
\end{enumerate}

This algorithm is called the error back propagation algorithm, and has been used for years to train multi-layer neural networks.

This means that we can legitimately consider each layer of our networks as a "black box", or a module, that takes some inputs and gives some outputs. We can forward data through it, and get the backpropagated gradient $\frac{\partial C}{\partial m^l}$ from it without actually knowing what it does or how it does it. This observation is at the origin of the modular architecture of the \texttt{Torch} framework that we used during our project and that is described later in chapter \ref{chap:torch}.

\subsection{Stochastic Gradient Descent}

As explained in section \ref{subsec:backprop}, we want to optimize an approximation of the risk over $N$ training samples. To do so, we first forward the whole dataset through the network, sum the gradients for each sample, and finally update all the weights of the network:

$$w_{t + 1} = w_t - \gamma \frac{1}{N} \sum_{i = 1}^N \frac{\partial C(x_i)}{\partial w_t}$$

where $w_t$ are the weights of the network at the current iteration, and $w_{t + 1}$ are the weights for the next iteration. In practice, however, we use a technique called stochastic gradient descent (see \cite{sgd} for more details). This technique consists in estimating the gradient of the whole dataset by the gradient of only one sample. This means that instead of considering all training examples before updating the weights, we do it after each of them:

$$w_{t + 1} = w_t - \gamma \frac{\partial C(x_i)}{\partial w_t}$$

This results in a faster convergence speed. Indeed, calculating the gradients on the whole dataset in the first iterations is a waste of time, because the network was randomly initialized and will thus certainly have a huge error on each sample. For this reason, we used stochastic gradient descent in all our experiments.

\subsection{Example: linear layer}

We will now give as example the calculations described in section \ref{subsec:backprop} for the case of a linear layer:

$$m^l(x) = W^lm^{l - 1}(x) + b^l$$

where $W^l \in \mathbb{R}^m \times \mathbb{R}^n$ and $b^l \in \mathbb{R}^m$. This linear layer has thus $n$ inputs and $m$ outputs. When applying the backpropagation algorithm, it receives $\frac{\partial C}{\partial m^l}$ from the next layer. We can rewrite the above equation as:

$$m^l(x) = (W^l_1 | W^l_2 \cdots | W^l_n) \left( \begin{array}{c}
m^{l - 1}_1(x) \\
m^{l - 1}_2(x) \\
\vdots \\
m^{l - 1}_n(x) \\
\end{array}\right) + \left( \begin{array}{c}
b^l_1 \\
b^l_2 \\
\vdots \\
b^l_m \\
\end{array}\right)$$

where $W^l_k$ are the columns of $W^l$. Thus, $m^l_j(x) = \sum_{k = 1}^n W^l_{jk}m^{l - 1}_k(x)$. We can then compute:

\begin{align*}
\frac{\partial C}{\partial W^l_{jk}} &= \frac{\partial C}{\partial m^l} \frac{\partial m^l}{\partial W^l_{jk}} \\
&= \frac{\partial C}{\partial m^l} x_k \\
\frac{\partial C}{\partial b^l_j} &= \frac{\partial C}{\partial m^l} \underbrace{\frac{\partial m^l}{\partial b^l_j}}_{= 1} \\
&= \frac{\partial C}{\partial m^l} \\
\frac{\partial C}{\partial m^{l - 1}_k} &= \sum_{j = 1}^m \frac{\partial C}{\partial m^l_j} \underbrace{\frac{\partial m^l_j}{\partial m^{l - 1}_k}}_{= W^l_{jk}} \\
&= \sum_{j = 1}^m \frac{\partial C}{\partial m^l_j} W^l_{jk} \\
&= \frac{\partial C}{\partial m^l} \cdot W^l_k \\
\end{align*}

\section{Transfer function}
\label{sec:theory-transfer-func}

A common and widely used transfer function for hidden neurons in a non-linear neural network is the hyperbolic tangent. In our project, following the work of Ronan Collobert, we chose to use a variant of this function, called the hard hyperbolic tangent, and defined as follows:

$$hardTanh(x) = \left\{ \begin{array}{lll}
1 & \text{if} & x > 1 \\
x & \text{if} & -1 < x < 1 \\
-1 & \text{if} & x < -1
\end{array}\right.$$

\begin{figure}
\begin{center}
\includegraphics[width=0.5\textwidth]{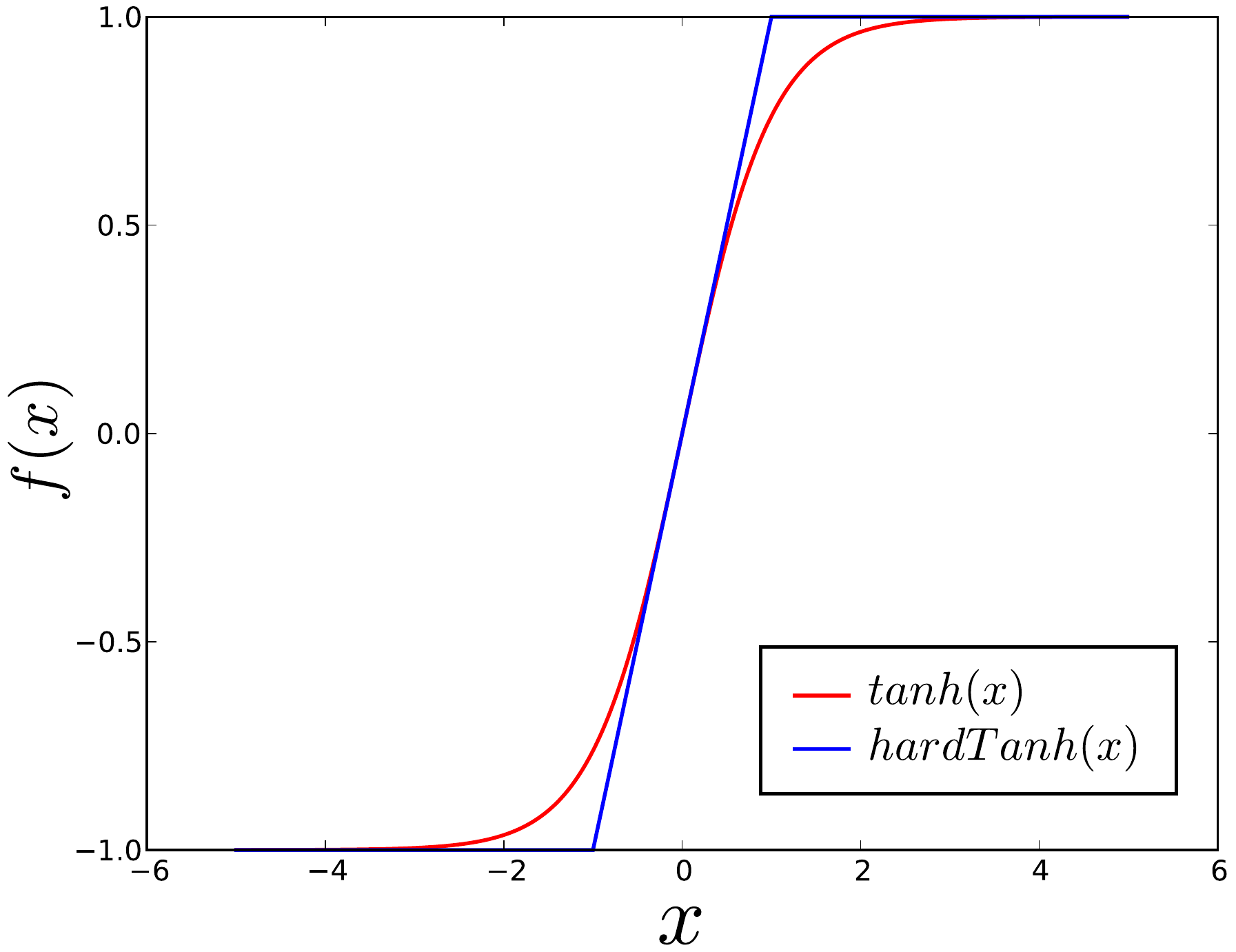}
\end{center}
\caption{Comparison of the $tanh(x)$ and $hardTanh(x)$ functions.}
\label{fig:transfer-functions}
\end{figure}

The two functions can be compared in figure \ref{fig:transfer-functions}. This substitution allows the hidden neurons to saturate faster, and thus puts the network in a non-linear mode sooner during the training process. Moreover, being much more simpler to implement than $tanh(x)$, $hardTanh(x)$ is faster to compute and thus makes our experiments run more quickly.

\section{Embedding}
\label{sec:theory-embedding}

When processing text using neural networks, one faces an important problem: these networks take real numbers as input, and not words. How could we convert the words into something usable by multi-layer perceptrons? We could simply replace each word by its index in a dictionary. While this approach would work, the main disadvantage is that it reduces greatly the information contained in the input: a simple number clearly cannot hold all the information contained in a word. Thus, we need a representation that still contains all the features of the original word, while being understandable and processable by a neural network.
\\

A solution to this problem was introduced by Yoshua Bengio in \cite{Bengio:2008}. In this article, he presents a way of mapping words to vectors in a $d$-dimensional space (typically, real-valued vectors in $\mathbb{R}^d$). The mapping from words to vectors is included in the neural network architecture, and the values of the vectors are learned during the training process. Collobert et al. refined this method in \cite{senna}.
\\

This technique is implemented as a module in \texttt{Torch}\footnote{See chapter \ref{chap:torch} for more information about the \texttt{Torch} framework.}, named \texttt{LookupTable}. It allows to learn and compute the mapping of $N$ different words into a $d$-dimensional space. In practice, getting the weight vector $w_i$ corresponding to a given word $i$ simply consists of a lookup in the table. However, it can be easily written as a matrix operation, to be later included in the matrix calculations that formalize a multi-layer perceptron, as explained in section \ref{sec:theory-nn}:
$$w_i = W * \left(\begin{array}{c} 0 \\ \vdots  \\ 0 \\ 1 \\ \vdots \\ 0 \\ \end{array} \right)
      \begin{array}{l} \\ \\ \\ \leftarrow i^{th} row \\ \\ \\ \end{array}$$

where $W$ is a $d \times N$ matrix which is multiplied by a very sparse vector of size $N$ that is one only on the $i^{th}$ row and $0$ everywhere else. Basically, the above formula extracts the $i^{th}$ column of the matrix $W$.
\\

The important fact that one should remember with these lookup tables is that the elements of $W$, i.e. the weights of the word vectors, are trained and learned jointly with the rest of the network !

\cleardoublepage
\chapter{Torch}
\label{chap:torch}

All the implementation part of this project was done using \texttt{Torch 5}\footnote{See \url{http://torch5.sourceforge.net} for more details}, an open-source machine learning framework written in C and Lua. This framework was originally developed by Ronan Collobert et al. (see \cite{torch}) at the IDIAP Research Institute in Switzerland. It provides a Matlab-like environment for state-of-the-art machine learning algorithms, with a very efficient implementation. Moreover, thanks to its modular architecture, it allows to easily add new algorithms or to change the behavior of any existing component. This chapter is a small introduction to this framework that gives some insight about its power and ease of use.

\section{Modular architecture}

The framework is divided in several packages, that each provides tools for specific tasks. The main package we used is \texttt{nn}\footnote{\url{http://torch5.sourceforge.net/manual/nn/index-1.html}}, which contains everything needed for building a feed-forward neural network and training it using backpropagation techniques. When using such multi-layer networks, each layer can be considered as a separate entity. Indeed, the only data we need when forwarding through a given layer is the output values of the previous layer. Similarly, when back-propagating the error through the network, all we need to update a given layer is the gradient values coming from the next layer, as explained in section \ref{subsec:backprop}.
\\

This observation allows to use each layer of the network as a \texttt{Module} (as called in the \texttt{Torch} framework). A \texttt{Module} has some generic methods used for forwarding data or backpropagating gradients. This allows to see them as "black boxes" and chain them, regardless of their content (they can be linear layers, convolutional layers, or even a complicated sequence of non-linear layers). The \texttt{nn} package provides containers for such modules that allow to build a sequence of several layers very easily, along with implementation of the most commonly used layers (linear layer, convolutional layers, etc.).nal layers, etc.).
\\

\texttt{Torch} also provides an efficient implementation of matrices of any dimension (called \texttt{Tensor}) and operators for performing several mathematical, statistical and manipulation operations on them. Finally, as each module implements forwarding and backwarding methods, it allows to train the created network very easily, using gradient descent technique. One can create a custom loss function, or use one of the criterions that are already implemented in \texttt{Torch} (mean-squared error, negative log-likelihood, etc.).

\section{Ease of use}

Figure \ref{torch-example} shows a code snippet describing how to create a simple non-linear multi-layer perceptron with three inputs, five hidden neurons with an hyperbolic tangent transfer function and one output. The resulting network is illustrated in figure \ref{torch-example-mlp}. This should give the reader an idea of how simple it is to word with neural network using \texttt{Torch}.

\begin{figure}
\begin{center}
\begin{lstlisting}
-- building mlp
mlp = nn.Sequential()
mlp:add( nn.Linear(3, 5) )
mlp:add( nn.Tanh() )
mlp:add( nn.Linear(5, 1) )

-- creating input
input = lab.random(3) -- random Tensor of size 3

-- forward input
output = mlp:forward( input )
\end{lstlisting}
\end{center}
\caption{Example of basic Torch usage: we first create a \texttt{Sequential}, which is a container that allows to build a sequence of layers. Then, we successively add a linear layer with three inputs and five outputs, a non-linearity, and finally a linear layer with five inputs and one output. The resulting network is pictured in figure \ref{torch-example-mlp}. After creating the network, we feed it with a random-valued input and get the corresponding output.}
\label{torch-example}
\end{figure}

\begin{figure}
\begin{center}
\includegraphics[width=0.5\linewidth]{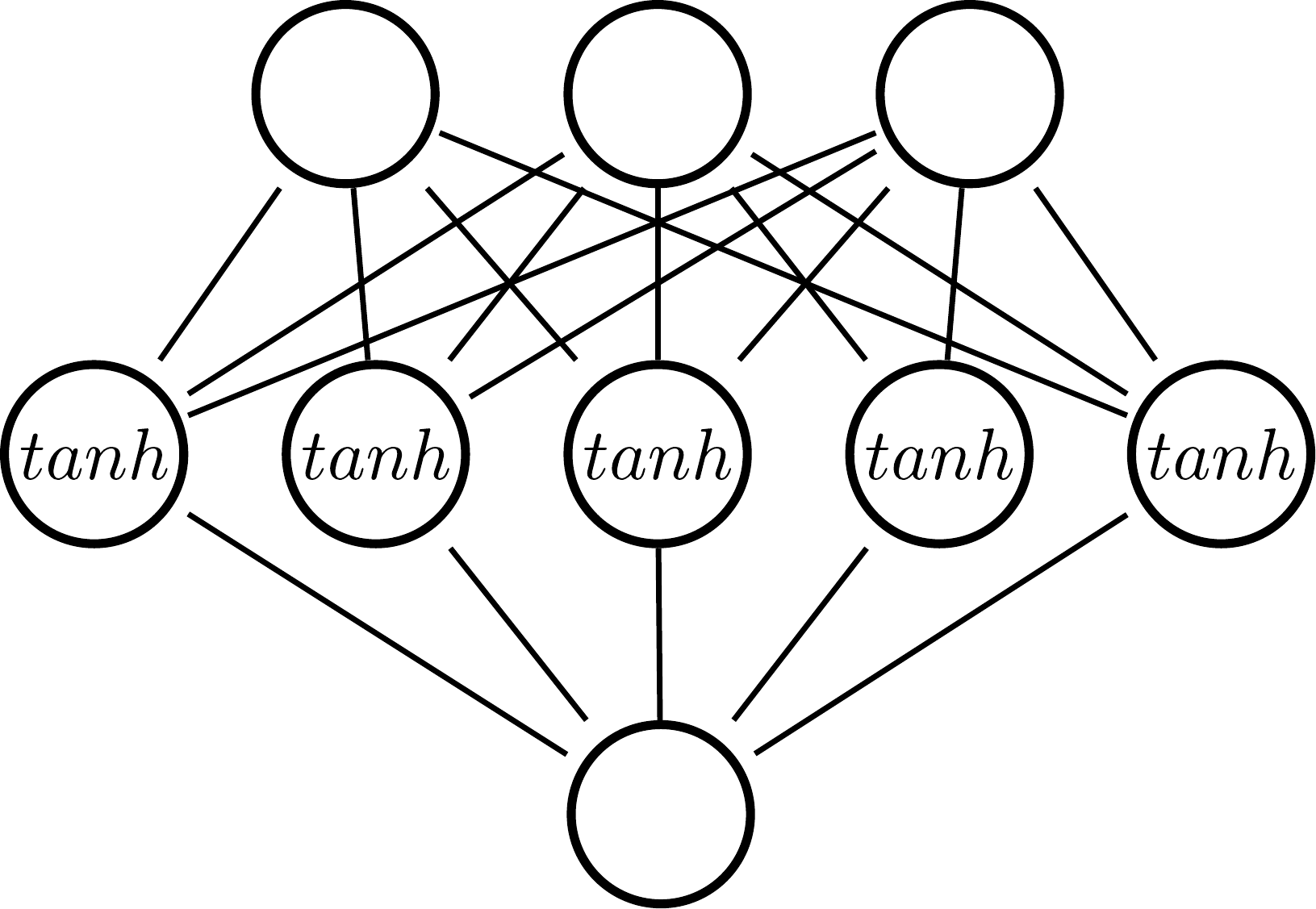}
\end{center}
\caption{Non-linear multi-layer perceptron corresponding to the code snippet in figure \ref{torch-example}, with tree inputs, five hidden neurons with an hyperbolic tangent transfer function and one output.}
\label{torch-example-mlp}
\end{figure}
\cleardoublepage
\chapter{Translation}
\label{chap:translation}

The first part of our project is about machine translation. This field of research has been of high interest for many years. With Internet, everybody has access to resources in several different languages. Moreover, with cheap air travels, it has become very easy to fly to foreign exotic countries, often with completely different language and even alphabet. Thus, the need for efficient translation tools is greater than ever.
\\

Different approaches exist to this problem, with more or less success. State-of-the-art translation tools, such as Google Translation\footnote{\url{http://translate.google.com}}, are mainly based on "mapping tables" : when given a sentence to translate, it looks into a huge database to find a sentence (or part of a sentence) for which it knows the translation. Then, it combines the known translations to build the final sentence. While this solution gives satisfying results, the main drawback is that it requires to possess the mapping table. Big companies like Google may have enough available data to build such database, and the infrastructure to store and use it, but it is clearly not feasible for individuals.
\\

Our solution to the translation problem is to train a neural network to embed words and sentences into a $d$-dimensional space. In this space, words or sentences having a close meaning will be close to each other, regardless of which language they belong to. This way, translating a word would be as easy as finding the closest word of the other language in the embedding space.

\section{Europarl Parallel Corpus}

To train such a task, we need a dataset of aligned sentences, i.e. pairs of sentences in two different languages, that are the translation of each other. There are not many of these available on the Internet, but fortunately some countries have several official languages, and thus translate all their official documents into each of them. This holds also for the European Union, which publishes its official documents in the language of each country that belongs to it. This is a great opportunity for machine translation, as it provides parallel datasets for translation to and from a lot of different languages, whereas the other available data is usually limited to a few common languages (e.g. English and Spanish).
\\

The Europarl Parallel Corpus\footnote{Available at \url{http://www.statmt.org/europarl/}} is extracted from the European Parliament proceedings, and contains versions in eleven different languages. It is distributed in a structured form, and also provides tools to generate sentence aligned data (see \cite{europarl} for more details). Such data consists of one text file per language, in which the $\text{i}^\text{th}$ line of each file correspond all to the same sentence, thus allowing to learn how to translate from one the others.
\\

As we needed to be able to assess the quality of the results during our experiments, we focused on English/French translations. However, our approach is language independent, and thus could be used with any languages for which aligned sentences are available.

\section{Using words only}

The first approach we used can be seen as a "bag of words" model: for each sentence, we consider its words separately, without taking into account their order. As described in the section \ref{sec:theory-embedding}, we embed the words into a $d$-dimensional space to process them. The weights of the vectors representing each word are randomly initialized, and trained during the learning process. To represent a sentence in this embedding space, we simply use the mean of all of its words.

\subsection{Network Architecture}

\begin{figure}
\begin{center}
\includegraphics[width=0.8\textwidth]{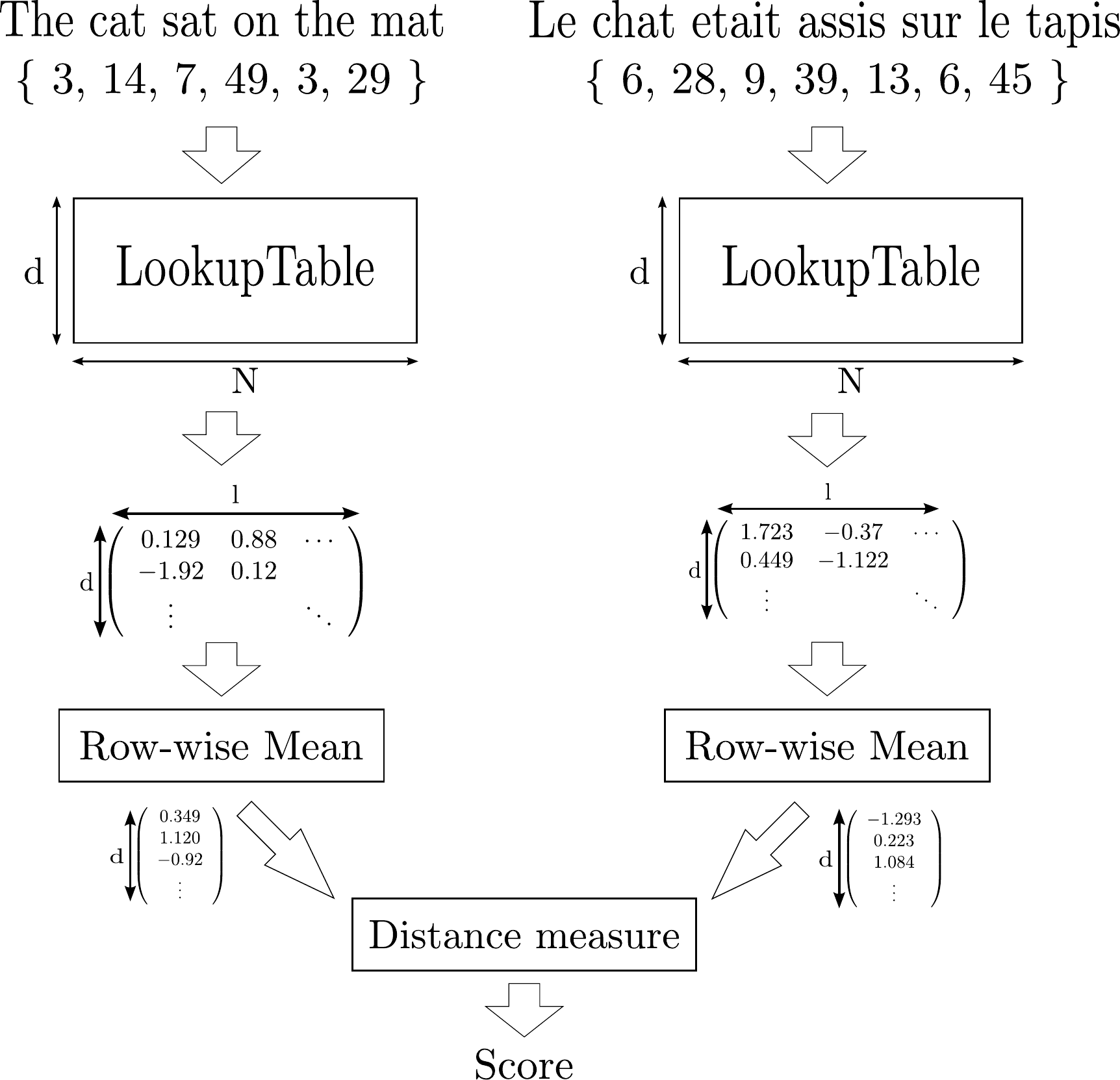}
\end{center}
\caption{Architecture of the multi-layer perceptron used to learn English/French translations. Two sentences, one in each language, are forwarded in parallel through a lookup table, to replace words by $d$-dimensional vectors. Then, we take the mean of the vectors of each sentence as their representation, and compute the distance between these two representations, to use it as a score.}
\label{fig:translation}
\end{figure}

Figure \ref{fig:translation} shows a diagram representing the architecture of the multi-layer perceptron we used to learn the translation task. As input, it receives a pair of two sentences, one in each language. Each word is represented by its index in a global dictionary, which has been computed offline and is constant for each language. The first stage of the network runs the list of indices through a lookup table, which replaces each word by its $d$-dimensional representation in the embedding space, as described in section \ref{sec:theory-embedding}. The output of the lookup table is thus a $d \times l$ matrix, where $d$ is the dimension of the embedding space and $l$ the length of the sentence.

Then, we take the mean of all its words as the representation of the whole sentence. These operations are done in parallel to the two sentences in each language. Finally, we measure the distance between the two sentences in the embedding space, and use this measure to assign a score to this particular pair of sentences (a short distance between the encoded representations means a high score, whereas a big distance means a low score).

\subsection{Training algorithm}

\begin{figure}
\begin{center}
\includegraphics[width=0.5\textwidth]{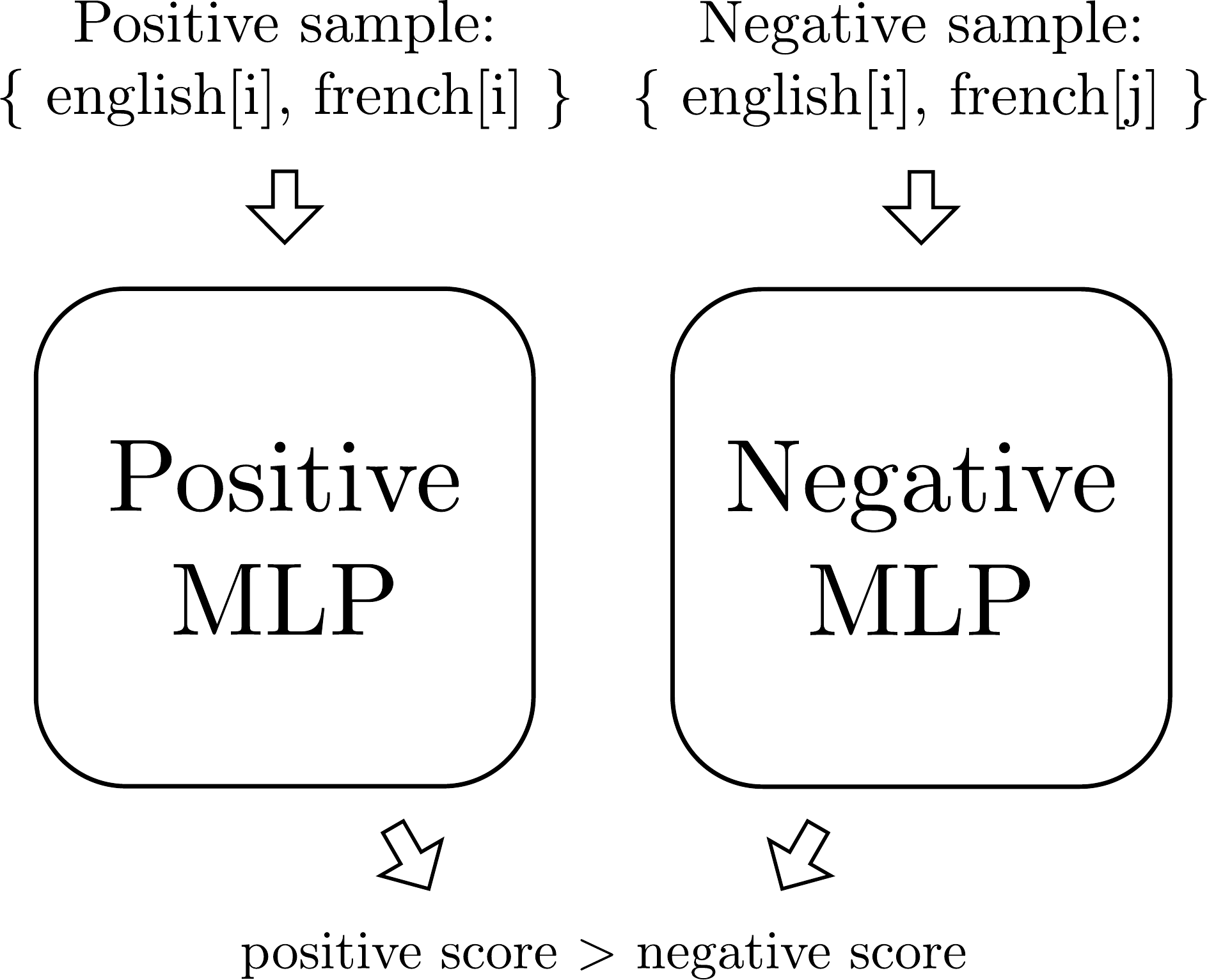}
\end{center}
\caption{Global architecture of the neural network. We take two instances of the network described in figure \ref{fig:translation} and feed them with one positive sample (two corresponding sentences) and a negative one (the corresponding French sentence as been replaced by a random one). We compare the two scores obtained as output, and want the score of the positive sample to be higher than the score of the negative one.}
\label{fig:translation-posNeg}
\end{figure}

The basic idea used for the training is that two sentences that are the translation of each other (we call such a pair "corresponding sentences") should have representations that are close to each other in the embedding space (i.e. a high score), whereas two random sentences should be far from each other (i.e. have a low score). To do so, we train a network composed of two copies of the network described above, as shown in figure \ref{fig:translation-posNeg}. For each line $i$ of the aligned corpus, we generate two samples: a positive one, that contains the $\text{i}^\text{th}$ English sentence and the $\text{i}^\text{th}$ French sentence, and a negative one, that contains the $\text{i}^\text{th}$ English sentence and a randomly chosen French sentence. We forward these two examples through the network and compare the resulting scores.
\\

\subsection{Loss function}

We use a margin ranking criterion as loss function. This criterion can be formalized as follows:

$$L_w(x) =max(0, m - f_w(x_{pos}) + f_w(x_{neg}))$$

where $x_{pos}$ and $x_{neg}$ are respectively the positive and negative samples described above, and $m$ is a fixed margin. This loss functions optimizes the network such that scores of positive samples are bigger than scores of negatives sample with a given margin. T
he value of the margin is arbitrary, the weights of the networks being dynamically adjusted. However, experiences have shown that the convergence speed is actually affected by this value. We thus empirically chose the square root of the dimension $d$ as a margin in our experiments.

\subsection{Performance evaluation}

To evaluate the performance of our network, a good metric would have been to compute the ranks of each pair of corresponding sentences and average them. The rank of a given pair of corresponding sentences \texttt{\{ english[i], french[i] \}} is defined as follows: it as a rank $k$ if \texttt{french[i]} is the $k^{th}$ closest French sentence to \texttt{english[i]}. Algorithm \ref{alg:rank} describes how this rank is actually computed.

\begin{algorithm}
\caption{Computes the rank of corresponding sentences $i$}
\begin{algorithmic}
\STATE distances $\leftarrow$ \{\}
\STATE codeEnglish $\leftarrow$ embed(english[i])
\FOR{$j = 1$ to $N$}
\STATE codeFrench $\leftarrow$ embed(french[j])
\STATE distance $\leftarrow$ computeDistance(codeEnglish, codeFrench)
\STATE distances[j] $\leftarrow$ \{ j, distance \}
\ENDFOR
\STATE sort(distances) \COMMENT{sorts the array on its second column in descending order}
\FOR{$j = 1$ to $N$}
\IF{distances[j][1] = i}
\RETURN i
\ENDIF
\ENDFOR
\end{algorithmic}
\label{alg:rank}
\end{algorithm}

If our translations were prefect, we would get an average rank of one, meaning that for each English sentence, the closest French sentence is the actual corresponding sentence. However, with 1'200'000 sentences in the train set and nearly 160'000 sentences in the test set, computing the real rank of each sentences pair would be quite time consuming, even if restricted to the test set only. Therefore, we needed an approximation of the actual rank that was fast to compute.
\\

We thus came with a method for statistically estimating the average rank. To do so, we use a limited number $M$ of randomly chosen French sentences instead of the whole dataset, and count how often the corresponding sentences have a shorter distance that the English sentence and one random French sentence. Algorithm \ref{alg:rank-est} describes this process more in details. We ran experiments with different values of $M$ and discovered that it was actually sufficient to compare the corresponding sentences to only one random pair to have a decent estimation of the performance, thanks to the great number of both train and test samples.

\begin{algorithm}
\caption{Estimation of the performance using $M$ random samples instead of the whole dataset}
\begin{algorithmic}
\STATE estimator $\leftarrow$ 0
\FOR{$i = 1$ to $N$}
\STATE codeEnglish $\leftarrow$ embed(english[i])
\STATE codeFrench $\leftarrow$ embed(french[i])
\STATE distance $\leftarrow$ computeDistance(codeEnglish, codeFrench)
\STATE allFurther $\leftarrow$ TRUE
\FOR{$j = 1$ to $M$}
\STATE codeRandom $\leftarrow$ embed(french[$randomNumber(1, N)$])
\STATE distanceRandom $\leftarrow$ computeDistance(codeEnglish, codeRandom)
\IF{distanceRandom $<$ distance}
\STATE allFurther $\leftarrow$ FALSE
\ENDIF
\ENDFOR
\IF{allFurther}
\STATE estimator $\leftarrow$ estimator + 1
\ENDIF
\ENDFOR
\RETURN estimator / N
\end{algorithmic}
\label{alg:rank-est}
\end{algorithm}

More formally, we defined the following indicator function:
$$\mathbf{1}_w(x_{pos}, x_{neg}) = 
\left\{\begin{array}{lll} 
1 & \text{if} & f_w(x_{pos}) > f_w(x_{neg}), \\
0 & \text{otherwise} &
\end{array}\right.$$

where $x_{pos}$ is a "positive" sample composed of two corresponding sentences \texttt{\{ english[i], french[i] \}}, $x_{neg}$ a "negative" sample \texttt{\{ english[i], french[j] \}} ($j$ being randomly chosen) and $f_w(x)$ computes the rank of a given pair of sentences. Then, our measure estimates the following:

$$P(w) = \mathbb{E}[\mathbf{1}_w(x_{pos}, x_{neg})]$$

\subsection{Results}

We first used the dot product as the measure of the similarity between two sentences:

$$d(x, y) = \sum_{i=1}^d x_iy_i$$

From the work of Collobert et al. in \cite{senna}, we started our experiments with a dimension of fifty for the embedding space. We also ran different experiments, with higher and lower dimensions, to compare the results. 
\\

\begin{figure}
\begin{center}
\includegraphics[width=1\textwidth]{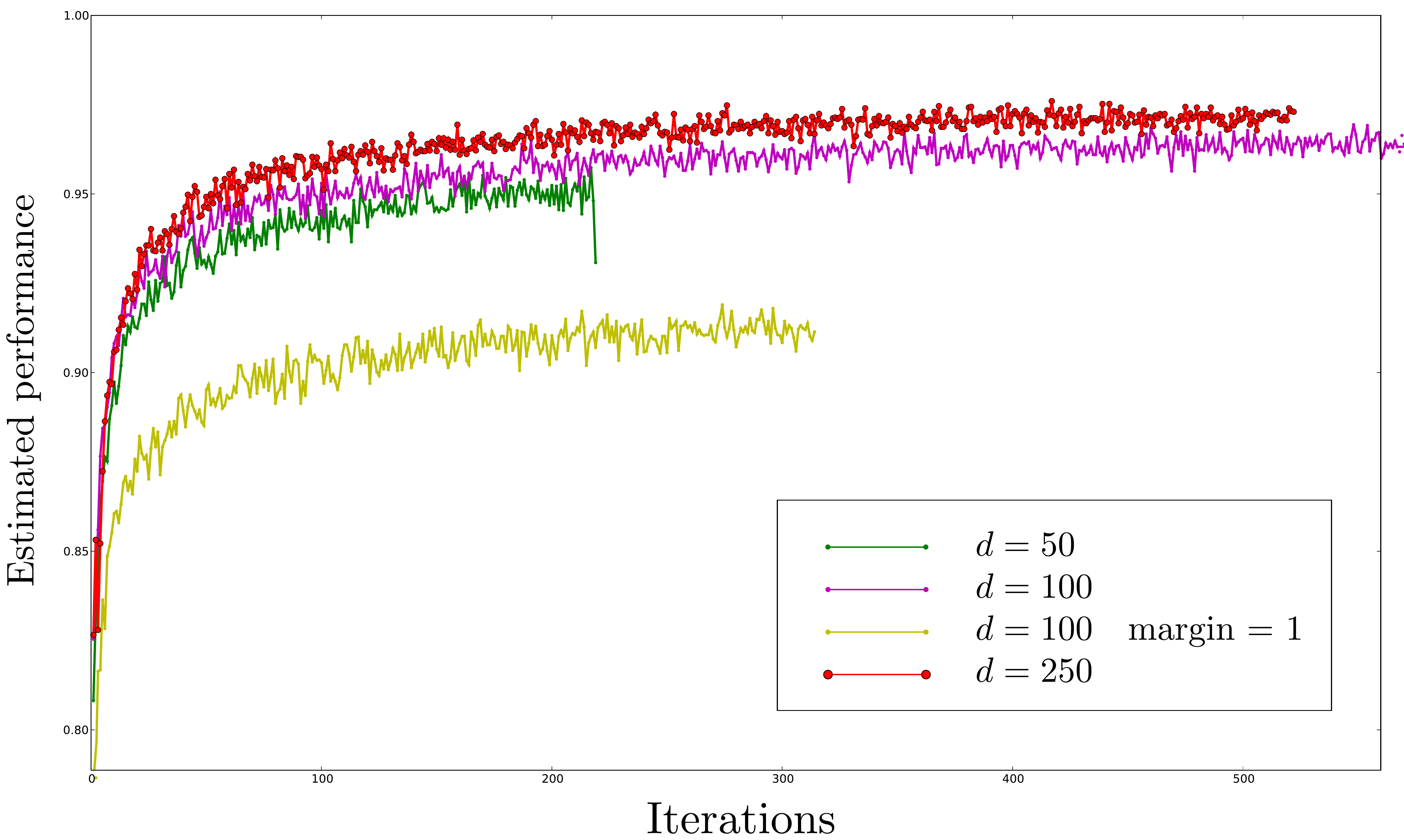}
\end{center}
\caption{Performance estimation of the network using words only and the dot product as measure of the distance, in function of the number of iterations, with several dimensions $d$. Interestingly, the yellow curve shows that using a margin of $1$ instead of $\sqrt{d}$ yields worse performances.}
\label{fig:translation-words-l2-results}
\end{figure}

The first results we obtained are shown in figure \ref{fig:translation-words-l2-results}. This plot displays the estimated performance evaluated on a test set of 10'000 sentences, with different dimensions $d$ of the embedding space. The best result obtained shows a performance of 95\%, which means that we would translate incorrectly a sentence 5\% of the time. While this may look like a good result, earlier experiments made by Jason Weston showed that an error rate as low as 0.5\% could be reached. After comparing our setups, we quickly found the difference between our experiments: he was using the L1 distance as measure, whereas we were using a dot product, which is equivalent to a L2 distance. As our measure gave more importance to great errors, some misaligned sentences in the data may have prevented our network to converge correctly.
\\

\begin{figure}
\begin{center}
\includegraphics[width=1\textwidth]{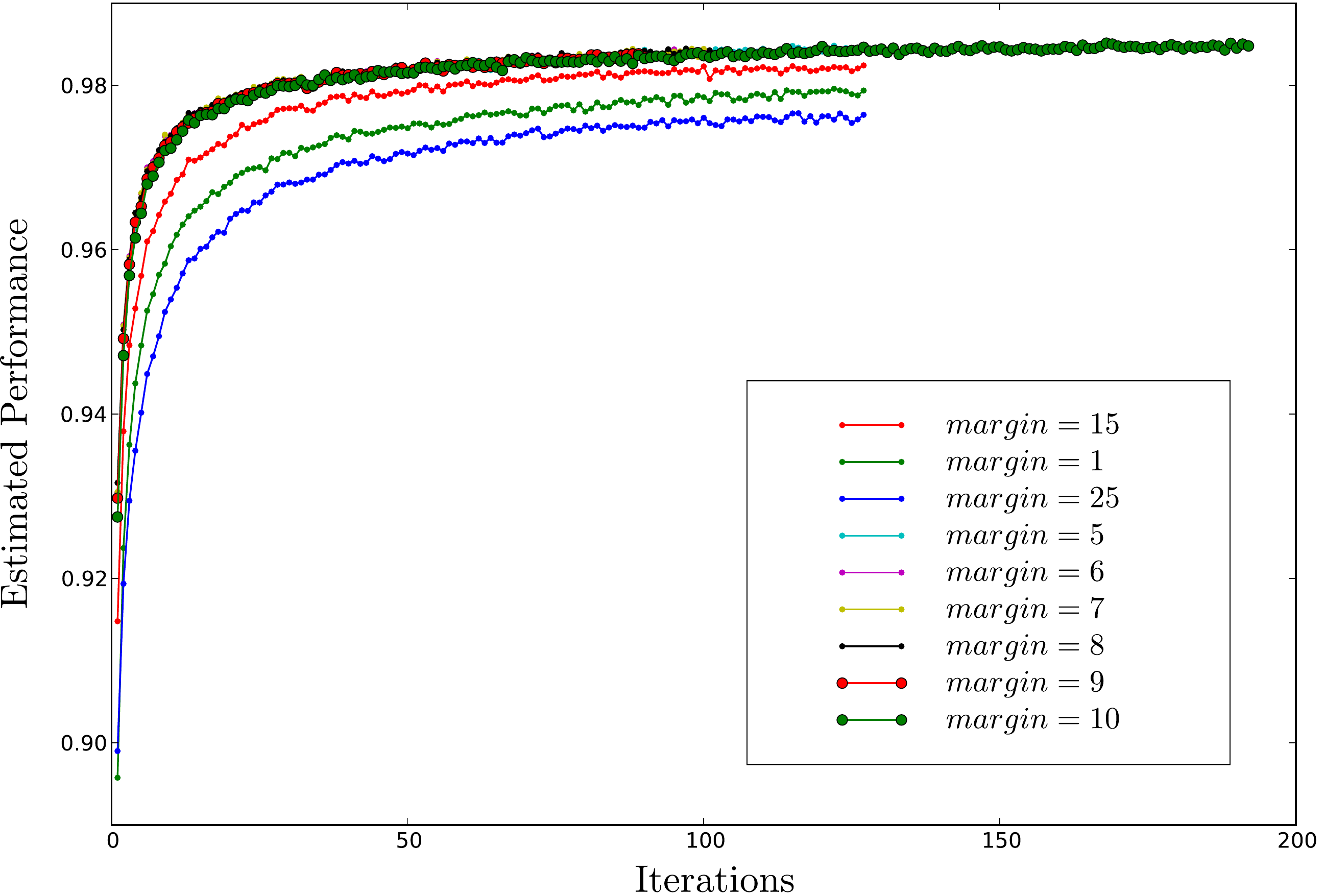}
\end{center}
\caption{Comparison of the performances of the network on words only, using several margins for the criterion, in function of the number of iterations. }
\label{fig:translation-margin}
\end{figure}

Figure \ref{fig:translation-words-l2-results} also showed something interesting: the experiments with a margin set to one instead of the square root of the dimension $d$ gave much worse results than the others. We did some more tests on the influence of the value of this margin. The results, presented in figure \ref{fig:translation-margin}, indicate that our empirical choice $\sqrt{d}$ as the margin of the criterion was good. While certainly still converging to the same result with other values, our margin allowed to converge faster.
\\

We changed our measure to the L1 distance, defined as follows, and ran again our experiments:

$$d(x, y) = \sum_{i=1}^d |x_i - y_i|$$

\begin{figure}
\begin{center}
\includegraphics[width=0.6\textwidth]{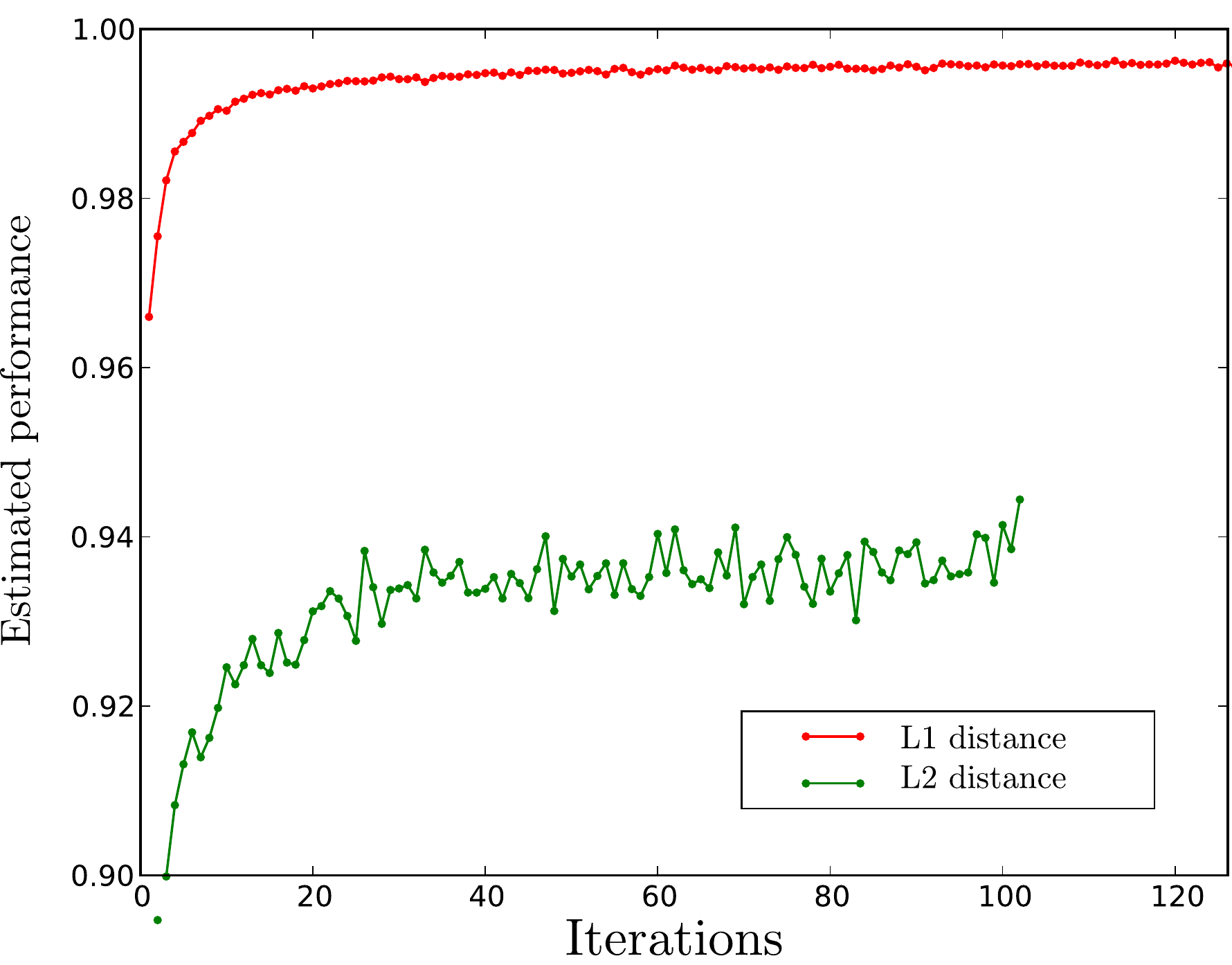}
\end{center}
\caption{Comparison of the performances using the L1 and L2 distances on words only, in function of the number of iterations, with an embedding dimension $d = 50$. }
\label{fig:translation-distances}
\end{figure}

This time, we obtained much more satisfying results, as illustrated in figure \ref{fig:translation-distances}. This figure shows the estimated performance on the test set of the same network, using a dimension $d = 50$ of the embedding space, but with the two different measures. While we obtained an error rate of 5.5\% with the dot product as measure of the distance, using the L1 distance instead allowed to reduce it to 0.03\%. After checking by hand which were the remaining errors, we discovered that they were inherent to the dataset. Indeed, it contains misaligned sentences, leading to samples that are impossible to translate correctly.
\\

\begin{figure}
\begin{center}
\includegraphics[width=1\textwidth]{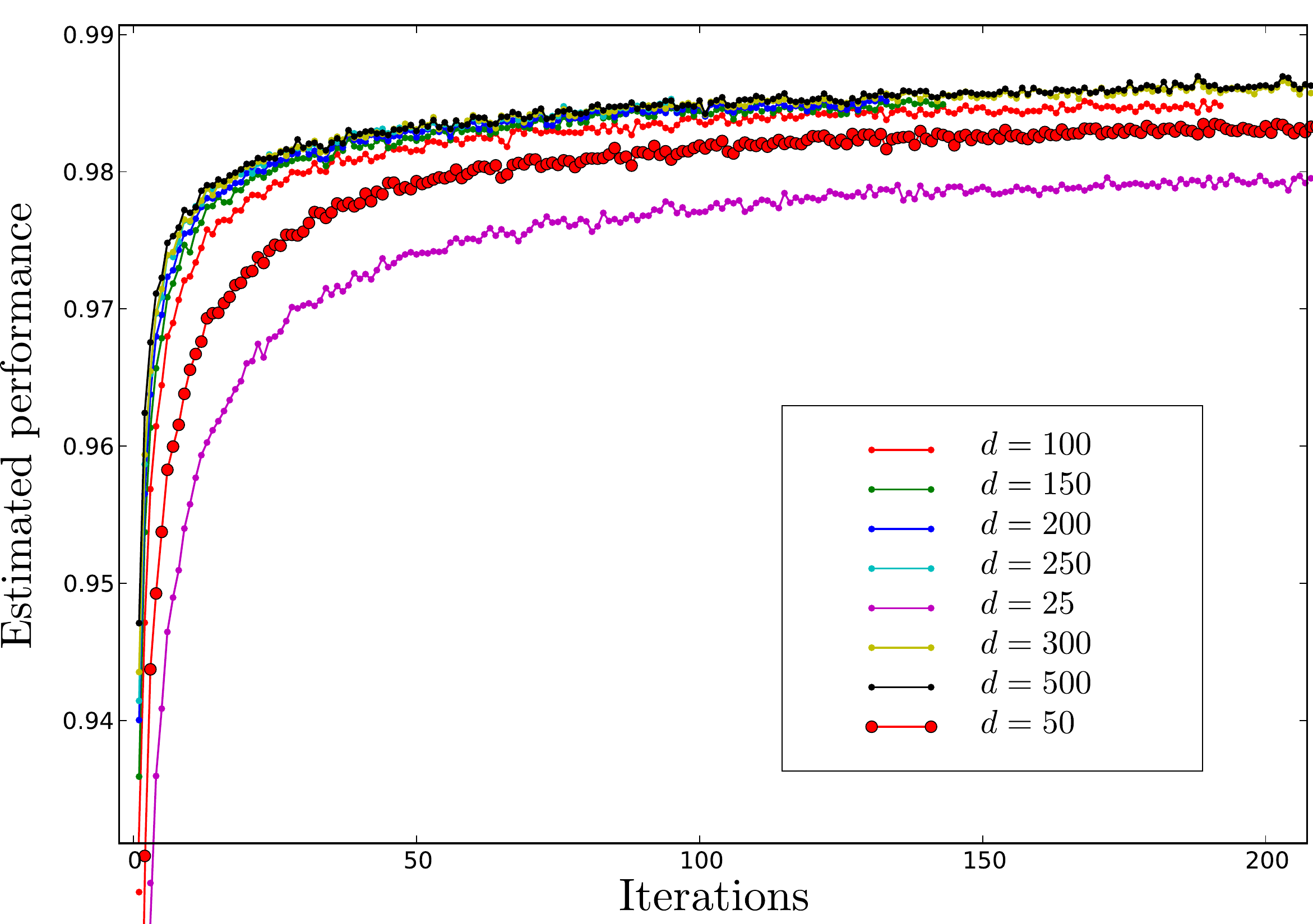}
\end{center}
\caption{Performance estimation of the network using words only and a L1 distance, in function of the number of iterations, with several dimensions $d$. }
\label{fig:translation-words-l1-results}
\end{figure}

We compared the results of the L1 distance with several dimensions $d$ in figure \ref{fig:translation-words-l1-results}, to see if we could reach better performances with higher embedding dimensions. While dimensions smaller than fifty clearly showed reduced performances, the results suggest that with dimensions of one hundred and above, the performances are not constrained by the lack of complexity of the network, but rather by our "bag of words" approach.
\\

Figure \ref{tab:translation-results-words} shows some English words and the closest French word in the embedding space, for a dimension $d = 50$. While these translations look correct, the really interesting result is obtained when looking at the other closest words in both languages, as pictured in figure \ref{tab:translation-results-closest-words}. We see that the embedding learned by the networks groups words that are similar, or related.

\begin{figure}
\begin{center}
\begin{tabular}{|l|l|}
\hline
european & europ\'{e}enne \\
commission & commission \\
president & pr\'{e}sident \\
between	& entre	 \\
against	& contre \\
human & humains \\
years & ann\'{e}es \\
great & grand \\
madam & madame \\
increase & augmenter \\
turkey & turquie \\
create & cr\'{e}er \\
sustainable & durable \\
\hline
\end{tabular}
\end{center}
\caption{Example of some English words and their closest French word in the embedding space.}
\label{tab:translation-results-words}
\end{figure}

\begin{figure}
\begin{center}
\begin{small}
\begin{tabular}{|l|l||l|l|}
\hline
european & europ\'{e}enne & commission & commission  \\
\hline
union & europ\'{e}ennes & committee & comit\'{e} \\
eu & europ\'{e}en & recommendations & recommandations \\
integration & union & council & commissions \\
europe & europ\'{e}ens & proposal & conseils  \\
citizenship & ue & advice & comit\'{e}s \\
europeans & institutions & recommendation & consultatif \\
enlarged & int\'{e}gration & document & conseil \\
constitution & \'{e}largie & consultation & document \\
unity & d\'{e}mocratique & dg & pr\'{e}sent\'{e}es \\
\hline
\hline
human & humains  & between & entre \\
\hline
rights & droits & links & relation  \\
 violations & homme & among & lien \\
minorities & violations & link & bilat\'{e}ral \\
religion & humain & relationship & bilat\'{e}rales \\
freedoms & humaine & amongst & liens \\
religious & religieuse & sides & diff\'{e}rences \\
crime & humaines & bilateral & ren\'{e}goci\'{e} \\
violence & dignit\'{e} & differences & bilat\'{e}raux \\
liberties & religion & stakeholders & diff\'{e}rentes \\
\hline
\hline
great & grand & turkey & turquie \\
\hline
tremendous & grande & cyprus & chypre \\
greatly & beaucoup & turkish & chypriote \\
enormous & grands & ukraine & turque \\
huge & \'{e}norme & belarus & turcs \\
immense & fortement & iraq & turc \\
extremely & \'{e}norm\'{e}ment & georgia & nations \\
greatest & immense & cypriot & turques \\
vast & consid\'{e}rables & democracy & r\'{e}f\'{e}rendums \\
considerable & extrêmement & membership & chypriotes \\
\hline
\end{tabular}
\end{small}
\end{center}
\caption{Sample of the embedding results: the header lines show English words and their closest French word in the embedding space. The other lines show the nine closest words in both languages to the corresponding English words on the header line.}
\label{tab:translation-results-closest-words}
\end{figure}

\section{Extending to n-grams}

Once we had obtained satisfying results considering words only, we extended our architecture to include more features in our learning process. Indeed, as explained above, using the mean of the sentence as its representation in the embedding space makes our approach similar to a "bag of words", in the sense that we lose the information about the order of the words, and their relation to each other. In order to address this problem and use as much information as possible in our solution, we added lookup tables for the pairs of words, trigrams, and so on. We could virtually include n-grams of any size, but did most of our testing using n-grams of length one and two (words and pairs of words).
\\

\begin{figure}
\begin{center}
\includegraphics[width=1\textwidth]{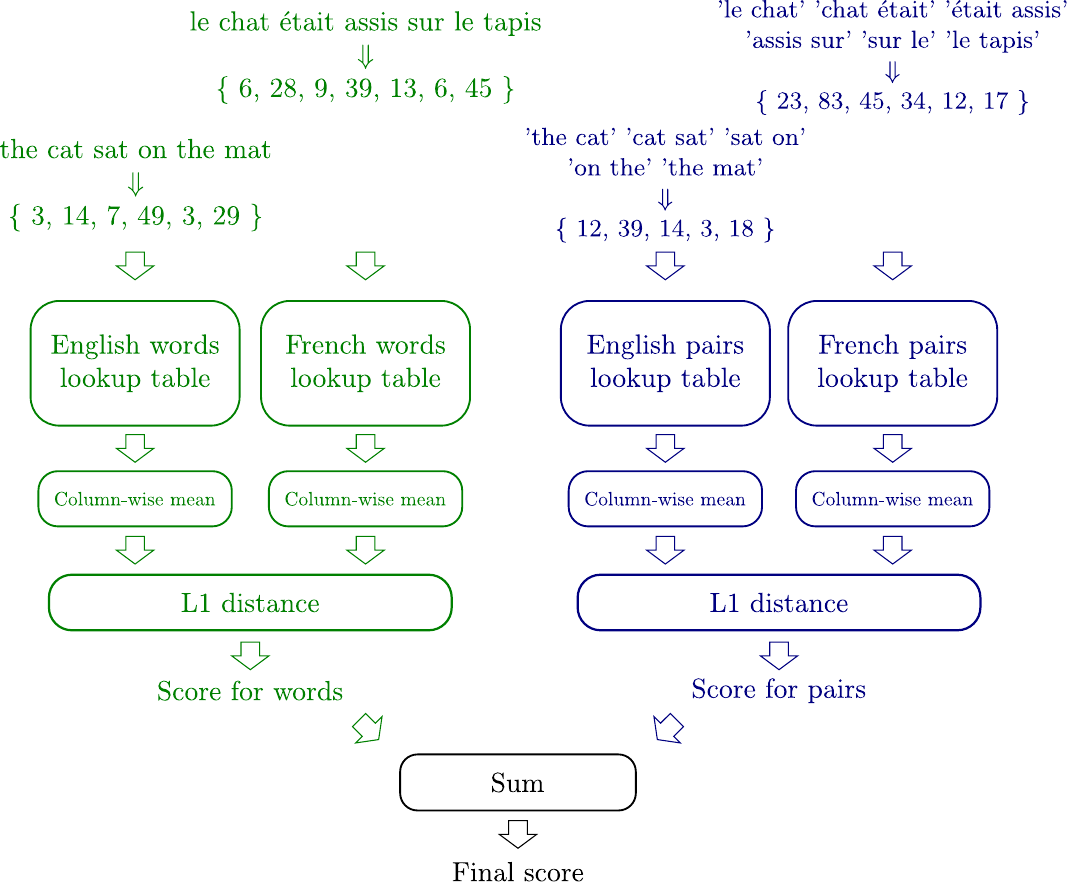}
\end{center}
\caption{Extended version of the network shown in figure \ref{fig:translation} to include pairs into the calculation of the score. N-grams of other sizes can obviously be included similarly.}
\label{fig:translation-mixed}
\end{figure}

\subsection{Network architecture}

Figure \ref{fig:translation-mixed} shows the modified architecture of the network that was presented in figure \ref{fig:translation}. We simply added a second network in parallel, this time using pairs of words instead of words only, with their own lookup tables. Each pair is replaced by its index in a dictionary, then mapped to its representation in the embedding space. As with words only, the mean of all these representations is then taken to get a representation of the sentence. This process is applied to both English and French sentences and the L1 distance between the two encoded representations is computed. Finally, we sum all these distances to obtain the global score of this particular pair of sentences, using words and pairs of words. We actually added more networks to train the translation from pairs to words (and vice-versa), and to learn the similarities between n-grams of different sizes in the same language. Note that we could easily take into account more n-grams by simply adding more of these networks in parallel, each having their own lookup tables for a given size of n-gram.
\\

The network is still trained with the same criterion as pictured in figure \ref{fig:translation-posNeg}: we have two copies of the network described above, and feed the first with corresponding sentences, while the other is given the same English sentence with a random French sentence. We want the score of the corresponding pair to be bigger than the score of the random pair. Moreover, to accelerate the training, we do not train all networks with each sample, but rather randomly select one network to train with each example between:

\begin{itemize}
\item all combinations of languages (English $\leftrightarrow$ English, English $\leftrightarrow$ French, French $\leftrightarrow$ French)
\item all combinations of sizes of n-grams (words $\leftrightarrow$ words, words$\leftrightarrow$ pairs, pairs $\leftrightarrow$ words, pairs $\leftrightarrow$ pairs)
\end{itemize}

For example, we may train the network to translate French pairs to English words at one step, and then train it to map English words to English pairs at the next step.

\subsection{Results}

\begin{figure}
\begin{center}
\includegraphics[width=1\textwidth]{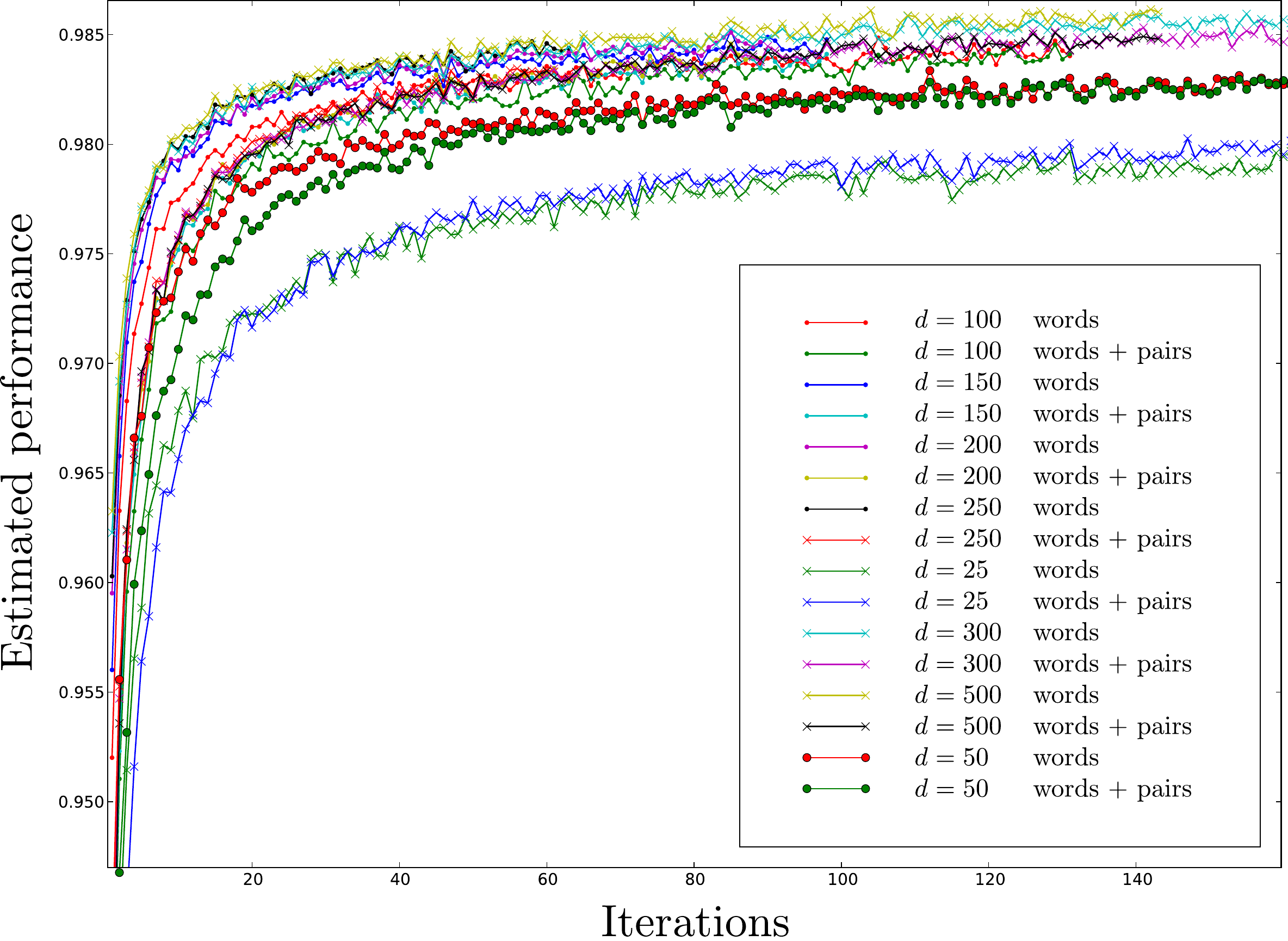}
\end{center}
\caption{Estimated performance of the extended network in function of the number of iterations, with different dimensions $d$ of the embedding space. For each experiment, we estimate the performance once using words only, then using words and pairs.}
\label{fig:translation-mixed-results}
\end{figure}

Figure \ref{fig:translation-mixed-results} shows the estimated performance of the extended version of the network, with different dimensions $d$ of the embedding space. For each dimension, we estimated the performance once using the words only, like we did with the simple version of the network, and then using both words and pairs of words to compute when computing the distance, as described above. We see that this second evaluation gives at first worse results than using words only, regardless of the dimension, certainly because it gives more information and thus needs more time for the network to adjust. After enough iterations, however, it gives the same results as using words only, and it shows even better performance afterwards (this is particularly visible on the curves of dimension $d = 25$, where the blue-crossed line starts under the green-crossed one, but finally gets over it). We also see that while having smaller dimensions $d$ of $25$ or $50$ is too limited and does not allow the network to fully represent the complexity of the problem, dimensions of 150 and higher have approximatively all the same performances.
\\

Again, when looking at the closest French words of a given English word, we obtain good translations, as seen with the previous networks. This also works with pairs of words, allowing us to directly translate English pairs to French pairs, and vice-versa. Moreover, we can look for results in the same language, i.e. see which English n-grams are close to a given English n-gram. Figure \ref{tab:translation-pairs-results} shows a few chosen English pairs and the closest English words. While some are simply variations of the two words of the pair, examples like "especially" being close to "in particular" makes us think that the network actually learned about the meaning of the words themselves, instead of simply building a English/French mapping.

\begin{figure}
\begin{small}
\begin{center}
\begin{tabular}{|l|l||l|l|}
\hline
of course & course & in particular & particular \\
\hline
course , & naturally & particular , & particularly \\
obviously , & obviously & , particularly & especially \\
but , & but & , especially & including \\
but it & nevertheless & particularly in & areas \\
nevertheless , & though & especially in & specifically \\
but there & certainly & , including & education \\
but that & however & and in & concerned \\
naturally , & well & concerned , & special \\
, but & nonetheless & areas , & regions \\
\hline \hline
public health & health & united states & united \\
health and & protection & the usa & usa \\
health . & environment & the us & america \\
consumer protection & safety & the united & states \\
the health & consumer & states of & nations \\
the protection & environmental & states , & country \\
of public & quality & the country & countries \\
protection of & public & states and & candidate \\
the environment & protecting & countries which & membership \\
protection . & competition & eu member & american \\
\hline
\end{tabular}
\end{center}
\end{small}
\caption{Samples of the embedding space learned by the extended network with a dimension $d = 150$. The header lines show chosen English pairs along with their closest English word, and the next lines show the other closest pairs and words to the chosen pairs.}
\label{tab:translation-pairs-results}
\end{figure}

\subsection{Discussion}

We are convinced that we could get even better results with this network simply by letting it train longer, as the results on the test set are clearly still increasing. However, while we obtained good translations for words and pairs of words, translating new sentences is far from obvious. Indeed, when presented with a new English sentence, we can embed it in the $d$ dimensional space using our neural network. But how could we actually find the corresponding French sentence? Clearly, we cannot list all possible sentences and pick the one that is the closest. We will present some ideas for solving this problem in chapter \ref{chap:conclusion}.
\\

At this point, we realized that while we learned good mappings from n-grams to n-grams, our approach was not so different then the ones presented in the introduction of this chapter. Our embeddings are compressed representations that captured some of the meaning of the sentences, but we felt that this was not enough. Indeed, it is more interesting to focus solely on learning the semantics of sentences, because this is a harder problem that would greatly help in a large number of fields, including machine translation. In fact, you have to understand what a sentence is about if you want to translate it correctly. We thus decided to end our work on translation and dive into the world of sentence compression.

\cleardoublepage
\chapter{Compression}
\label{chap:compression}

In this chapter, we present the second part of our work that focused on compressing sentences using neural networks. As explained in the end the previous chapter, our work on translation can be related to compression: we were taking sentences of arbitrary length and mapping them to a features vector of dimension $d$, to later compare these vectors between sentences. Basically, this features vector can be seen as a compressed representation of a sentence. However, because of the "bag of words" approach we used by taking the mean of all the words of the sentence, we lost a lot of information.

\section{General Idea}

We thus decided to try a new approach, still based on the idea of embedding words (and sentences) into a $d$-dimensional space, but this time with a different purpose than translation: semantic compression. We want to have an encoded representation that gives information about the meaning of the sentence. With the huge amount of data available everywhere on the web, we are more than ever in need of solutions for indexing, sorting, grouping and comparing text documents. However, comparing two texts is difficult: one could be short and summarized and the other long and verbose, while still talking about the same subject. Comparing word occurrences and extract keywords may help to find similarities between texts, but would certainly fail when comparing a complicated article of Wikipedia in English and its sibling in Simple English\footnote{\url{http://simple.wikipedia.org} - Articles in the Simple English Wikipedia use fewer words and easier grammar than the Ordinary English Wikipedia.}. Having a simple way of producing an encoded representation of sentences (and then whole documents) that synthesizes their meaning and semantics would help greatly in that direction. Instead of trying to compare text, we would work with points of a $d$-dimensional space, allowing us to use well known tools and algorithms for clustering, sorting, and so on.

\subsection{Compression}

\begin{figure}
\begin{center}
\includegraphics[width=0.7\textwidth]{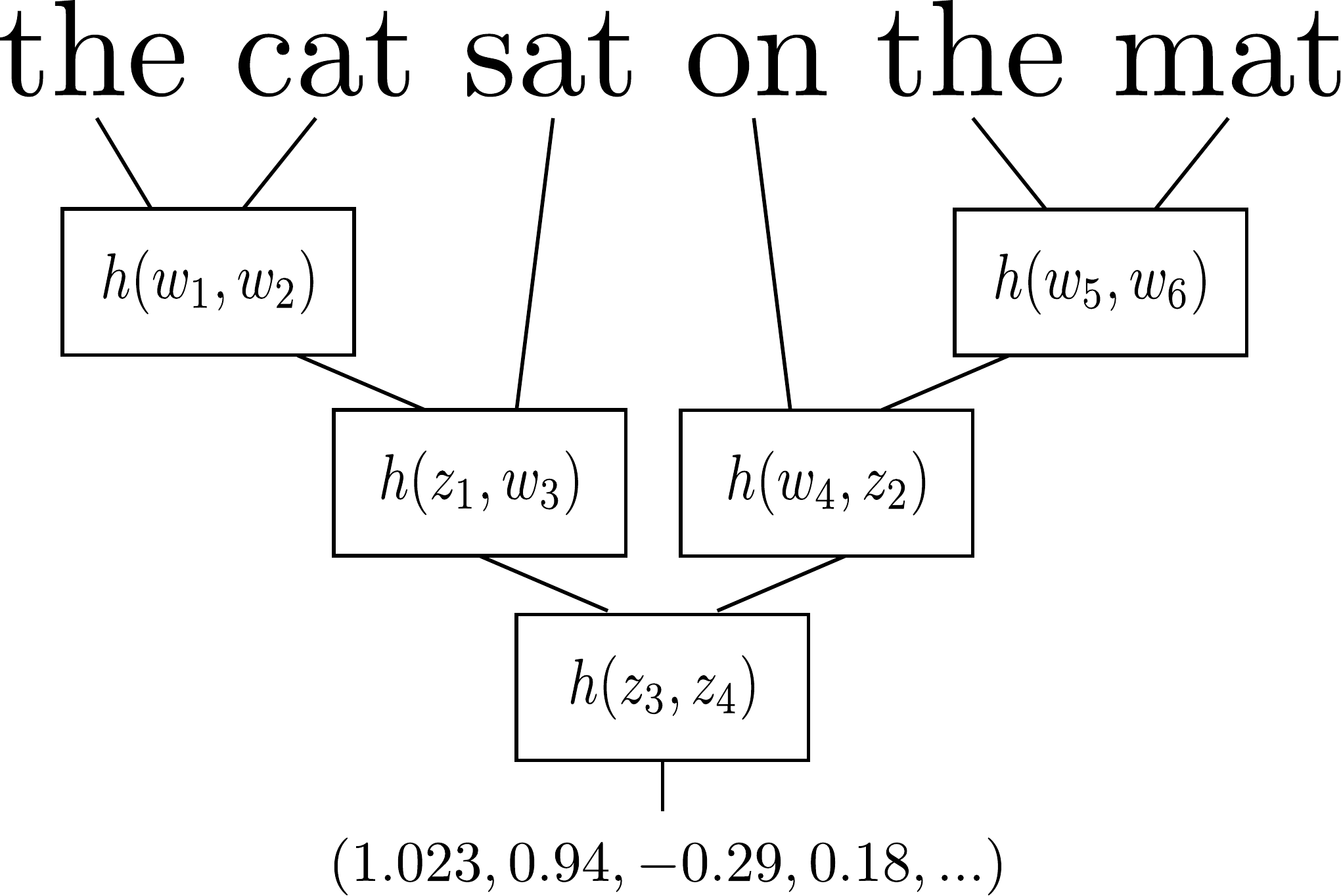}
\end{center}
\caption{Illustration of a sentence compression. Using a compression function $h(x,y) : \mathbb{R}^d \times \mathbb{R}^d \rightarrow \mathbb{R}^d$ that compresses two elements of the embedding space to another element of this same space, we iteratively group the words of a sentence two by two until we only have one element left. The order in which we group the elements can vary: in this example, we used the tree shown in the upper-left part of figure \ref{fig:compression-trees} but could have used any other.}
\label{fig:compression-idea}
\end{figure}

Our idea is thus to find a compression function $h(x,y) : \mathbb{R}^d\times\mathbb{R}^d \rightarrow \mathbb{R}^d$ that compresses two elements of a $d$-dimensional space to another element of this space. Then, we would apply this function iteratively to a sentence to reduce it to a single element of the embedding space, and use this vector as the encoded representation of the sentence. An example of this process is shown in figure \ref{fig:compression-idea}. There are therefore two main aspects to this problem:

\begin{enumerate}
\item find a good candidate for the compression function $h(x,y)$
\item determine the optimal order in which to group the elements two by two
\end{enumerate}

\begin{figure}
\begin{center}
\includegraphics[width=0.7\textwidth]{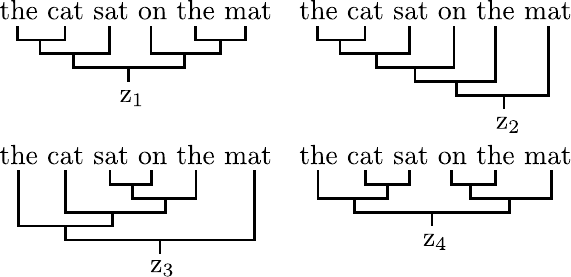}
\end{center}
\caption{Illustration of four different ways of grouping the words of a sentence. We can see that a sequence of groupings can be represented as a binary tree. We call the upper right tree a "left to right" tree, as we always group the two leftmost elements.}
\label{fig:compression-trees}
\end{figure}

Indeed, as shown in figure \ref{fig:compression-trees}, there are many ways of grouping a sentence two by two\footnote{In fact, we can see the sequence of groupings as a binary tree, and will later refer to it as such.}. The quality of the resulting representation may depend on the order in which the function $h$ is applied: maybe grouping first words that are related (determinants with nouns, adjectives with nouns) is better than grouping the sentence randomly, or with a left to right approach? However, finding both the function $h$ and the best way of grouping words at the same time would be hard. Therefore, we first used a fixed "left to right" tree (as pictured in the upper-right example of figure \ref{fig:compression-trees}) while learning $h$, and then used the resulting function $h$ to try and determine the optimal tree for any sentence.

\subsection{Extraction}

\begin{figure}
\begin{center}
\includegraphics[width=0.7\textwidth]{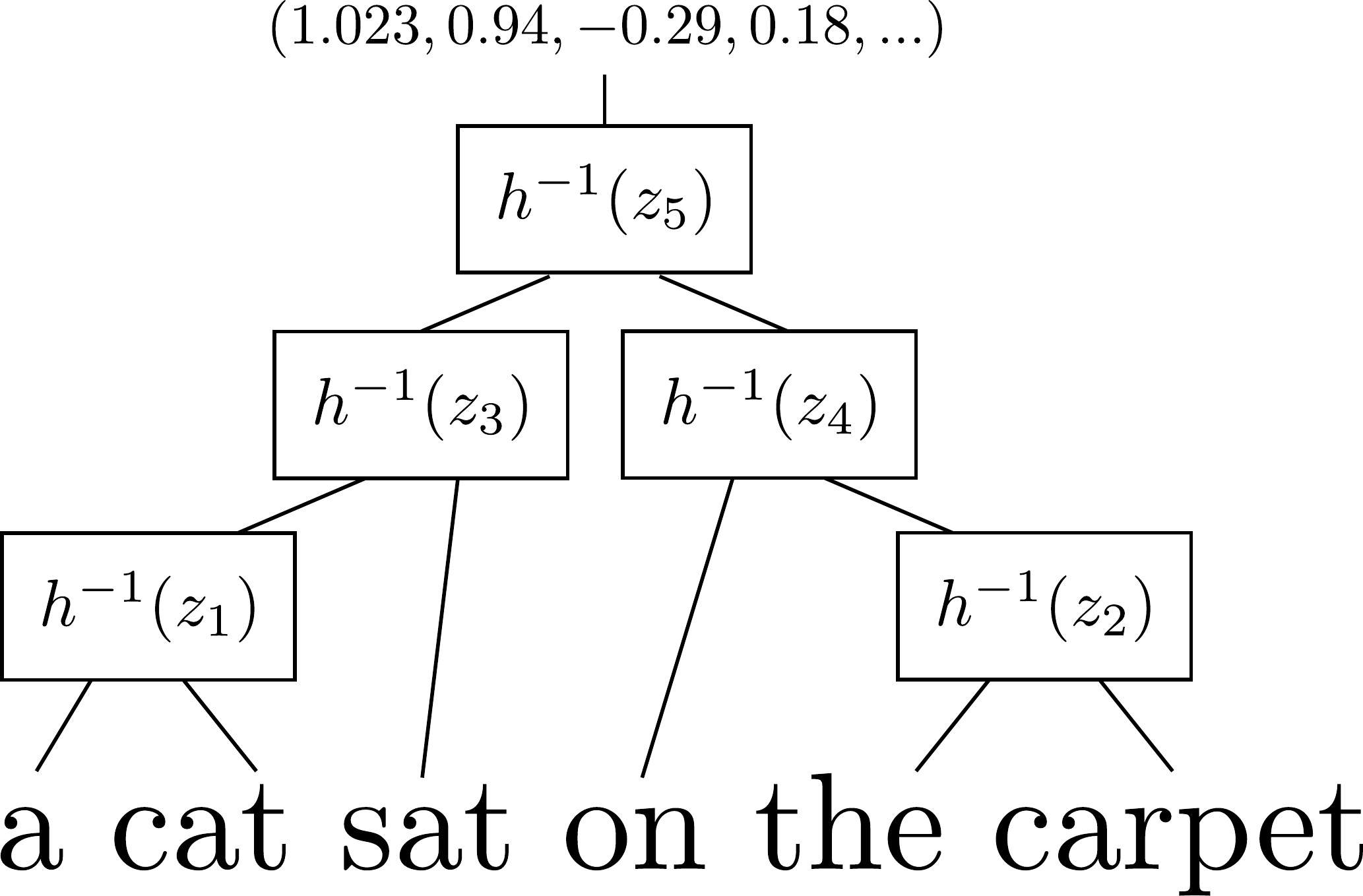}
\end{center}
\caption{Illustration of a sentence extraction. We iteratively apply the inverse of the compression function $h^{-1}(z) : \mathbb{R}^d \rightarrow \mathbb{R}^d \times \mathbb{R}^d$  to the encoded representation of the sentence in the exact opposite order as $h$ was applied during the compression phase. The resulting sentence may not exactly be the same as the original one, but should have a similar meaning.}
\label{fig:extraction-idea}
\end{figure}

Once we have a compressed representation of a sentence, we could optionally want to extract the sentence back. While this is not directly needed for comparing and clustering text information, it could be interesting to use this scheme as any other compression algorithm, i.e. to reduce the size of the data. To extract the information back from the encoded representation, we need the inverse of the compression function $h^{-1}(z) : \mathbb{R}^d \rightarrow \mathbb{R}^d \times \mathbb{R}^d$. This function can be trained jointly with $h$, or separately. We also need to know in which order the elements of the sentence were grouped, such that we could iteratively apply $h^{-1}$ in the exact opposite order to finally obtain the original sentence back. The extraction process is illustrated in figure \ref{fig:extraction-idea}. The compression may however not be lossless, but as the semantics are encoded in the compressed representation, the meaning of the extracted sentence should hopefully be close to the original one.

\section{Dataset}

For these experiments, we used the English content of Wikipedia as a dataset. This dataset was built by downloading a dump of English Wikipedia\footnote{available at \url{http://download.wikimedia.org/enwiki/}}, extracting and cleaning all paragraphs (removing any non-text data, converting all words to lower case, etc.) and finally shuffling them. The result is a random list of more than nine millions of paragraphs, totalizing more than six hundred ninety millions of words. From this dataset, we built a dictionary of all the different words (about two and a half millions) and sorted them by frequency. Finally, we thresholded these words to keep only the $N$ most frequent\footnote{On more than 2.5 millions words, only 100'000 of them appear 100 times or more in the whole wikipedia dataset. Most of the words are thus rare words, and are not worth considering during the training of the network.} (typically, $N = 1000$, $5000$ or $30'000$) and discarded all other words while training on the dataset.
\\

As we do not have sentences delimitation in this dataset, we cannot use actual sentences during our training. Instead, we slide a window of a fixed size over each paragraph, and extract all n-grams that do not contain a thresholded word. This means that we can obtains n-grams that are split between two actual sentences (e.g. "on the mat . The dog was"). Ideally, we should not consider these, as they make less sense than an actual sentence, but we are willing to pay the price of having less than perfect data in exchange to having a very large quantity of it.

\section{First approach: Auto-encoder networks}
\label{sec:compression-auto-encoder}

Our first approach was to use an auto-encoder network to learn both $h$ and $h^{-1}$ at the same time, and them use the two compression and extraction parts separately.

\subsection{Auto-encoder networks}

\begin{figure}
\begin{center}
\includegraphics[width=0.5\linewidth]{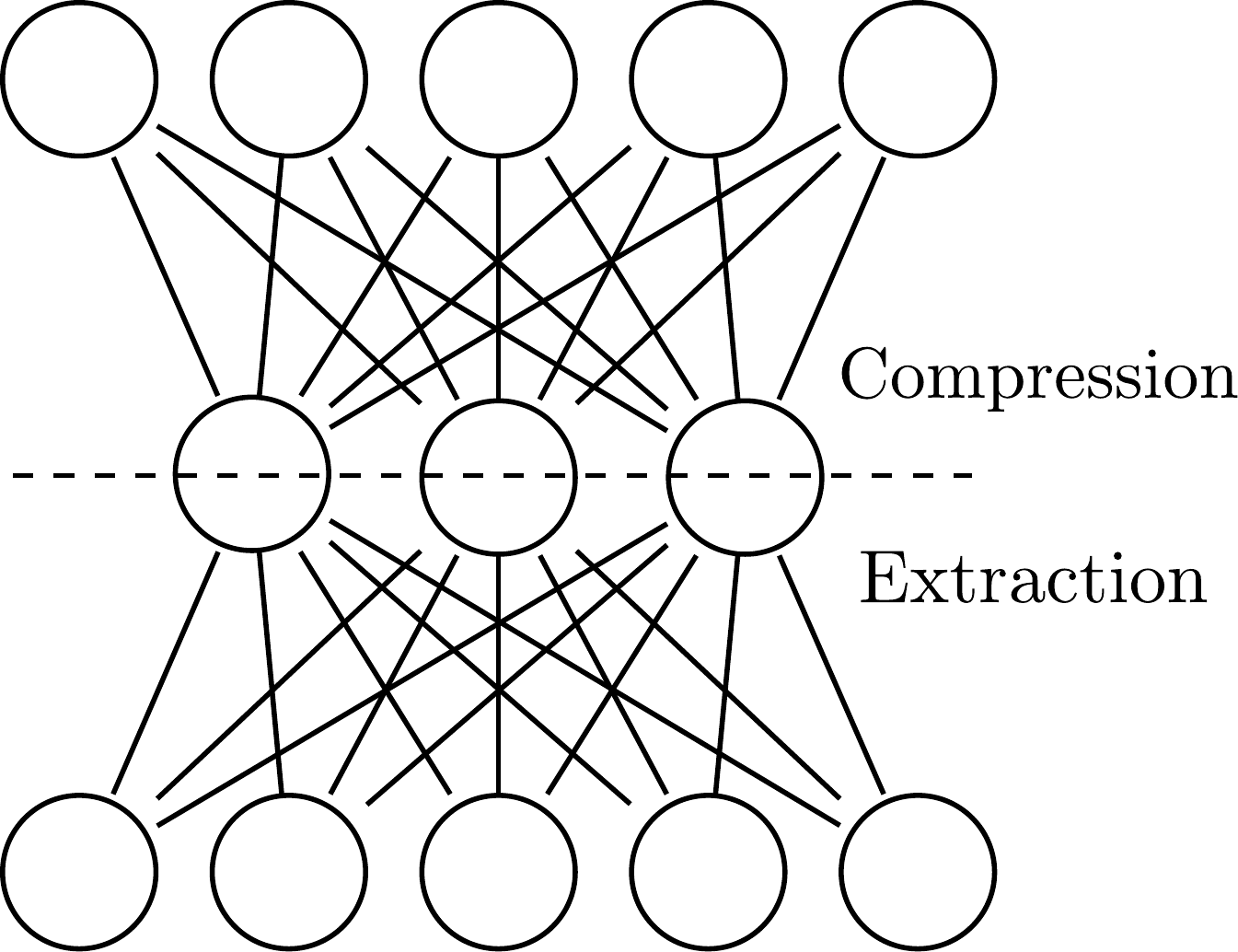}
\end{center}
\caption{Example of an auto-encoder network. Such network is usually symmetric and has the same number $n$ of inputs and outputs, along with a number $m < n$ of hidden neurons.}
\label{fig:auto-encoder}
\end{figure}

An auto-encoder network (also called autoassociator) is a special kind of multi-layer perceptron, that is usually symmetric. As illustrated in figure \ref{fig:auto-encoder}, the main features of such a network is that it has:

\begin{enumerate}
\item the same number $n$ of inputs and outputs
\item a number $m < n$ of hidden neurons
\end{enumerate}

The network is trained, using regular backpropagation techniques, to have its output equal to the input data (this technique has several names: identity mapping, encoding or auto-association). Once the training has converged, we can split the network in two parts, as illustrated in figure \ref{fig:auto-encoder}: 

\begin{itemize}
\item the compression part, composed of the upper part of the network
\item the extraction part, composed of the lower part of the network
\end{itemize}

By taking the output of the compression part, we get a compressed (encoded) representation of the input. Then, we can use the extraction part to get the original data back from its encoded representation. With a linear architecture similar to the one pictured in figure \ref{fig:auto-encoder}, Bourlard et al. have shown in \cite{Bourlard_auto} that an auto-encoder network is equivalent to applying Principal Component Analysis\footnote{Principal component analysis (PCA) involves a mathematical procedure that transforms a number of possibly correlated variables into a smaller number of uncorrelated variables called principal components. For more details, see \cite{wiki-pca}.} to the data. This means that instead of training the network, we could simply express the solution of the problem explicitly in terms of the input data, using techniques like SVD\footnote{Singular Value Decomposition}. This way, we would be sure to get the optimal solution, whereas a classical error backpropagation algorithm could get stuck in a local minima.
\\

It is also possible to add more hidden layers before and after the encoded representation, this time with non-linear transfer functions like the hyperbolic tangent. It as been a common misperception in the Neural Network community that even with non-linearities, auto-encoder networks trained with backpropagation are equivalent to linear methods such as PCA, as claimed by \cite{Bourlard_auto}. Japkowicz et al. have shown in \cite{Japkowicz00nonlinearautoassociation} that this is not the case: non-linear autoassociators behave actually differently from linear methods, can outperform them in projection or classification tasks and are even able to perform non-linear classification.

\subsection{Network architecture}

As explained in section \ref{sec:theory-embedding} and in the previous chapter, working directly with text is not really convenient. Hence, we replaced each word with its index in the dictionary. Then, when considering a sentence $s = \{ x_1, x_2, ..., x_l \}, x_i \in \{1, N\}$, we first ran it through a lookup table that maps each index to a vector in the embedding space. We obtained the matrix $S = (w_1 | w_2 ... | w_l), w_i \in \mathbb{R}^d$. Then, we could iteratively apply $h$ to any two columns of this matrix, regardless if they were word vectors from the lookup table or results of previous applications of $h$.
\\

\begin{figure}
\begin{center}
\includegraphics[width=0.7\textwidth]{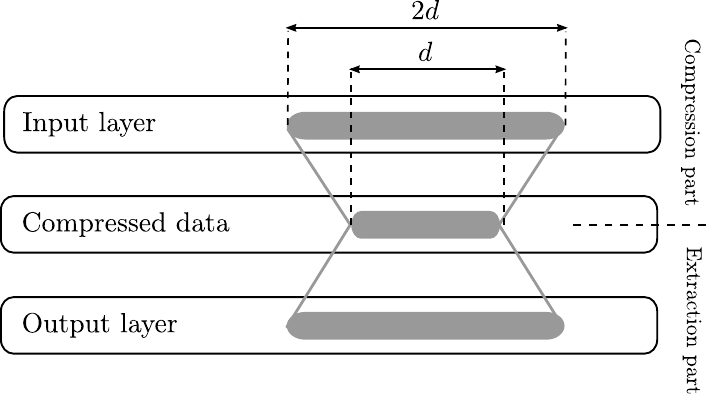}
\end{center}
\caption{Implementation of the compression function $h$ and extraction function $h^{-1}$ as a linear auto-encoder network with $2d$ inputs/outputs and $d$ compressed units.}
\label{fig:compression-linear}
\end{figure}

\begin{figure}
\begin{center}
\includegraphics[width=0.7\textwidth]{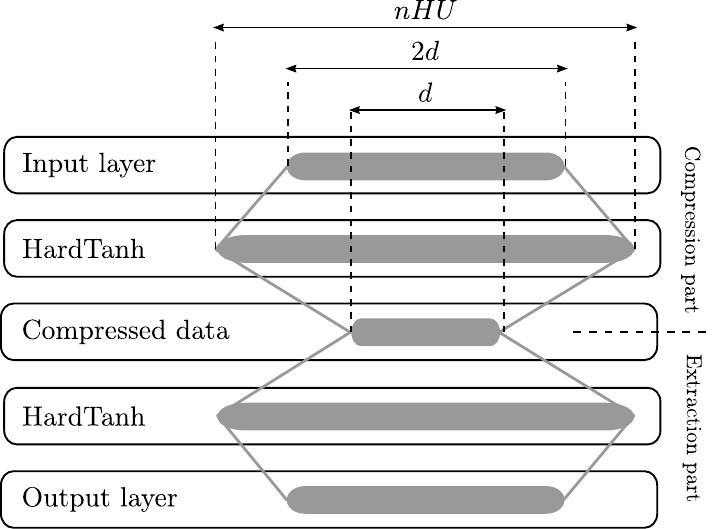}
\end{center}
\caption{Implementation of the compression function $h$ and extraction function $h^{-1}$ as a non-linear auto-encoder network with $2d$ inputs/outputs, $nHU$ hidden units and $d$ compressed units.}
\label{fig:compression-non-linear}
\end{figure}

The function $h$ can simply be implemented as the compression part of an auto-encoder network, and $h^{-1}$ as the extraction part of the same network, as illustrated in figure \ref{fig:compression-linear}. We tried both linear and non-linear setups, the non-linear simply adding a hidden layer with $nHU$ hidden units and a non-linear transfer function as shown in figure \ref{fig:compression-non-linear}.

\subsection{Local optimization}

\begin{figure}
\begin{center}
\includegraphics[width=1\textwidth]{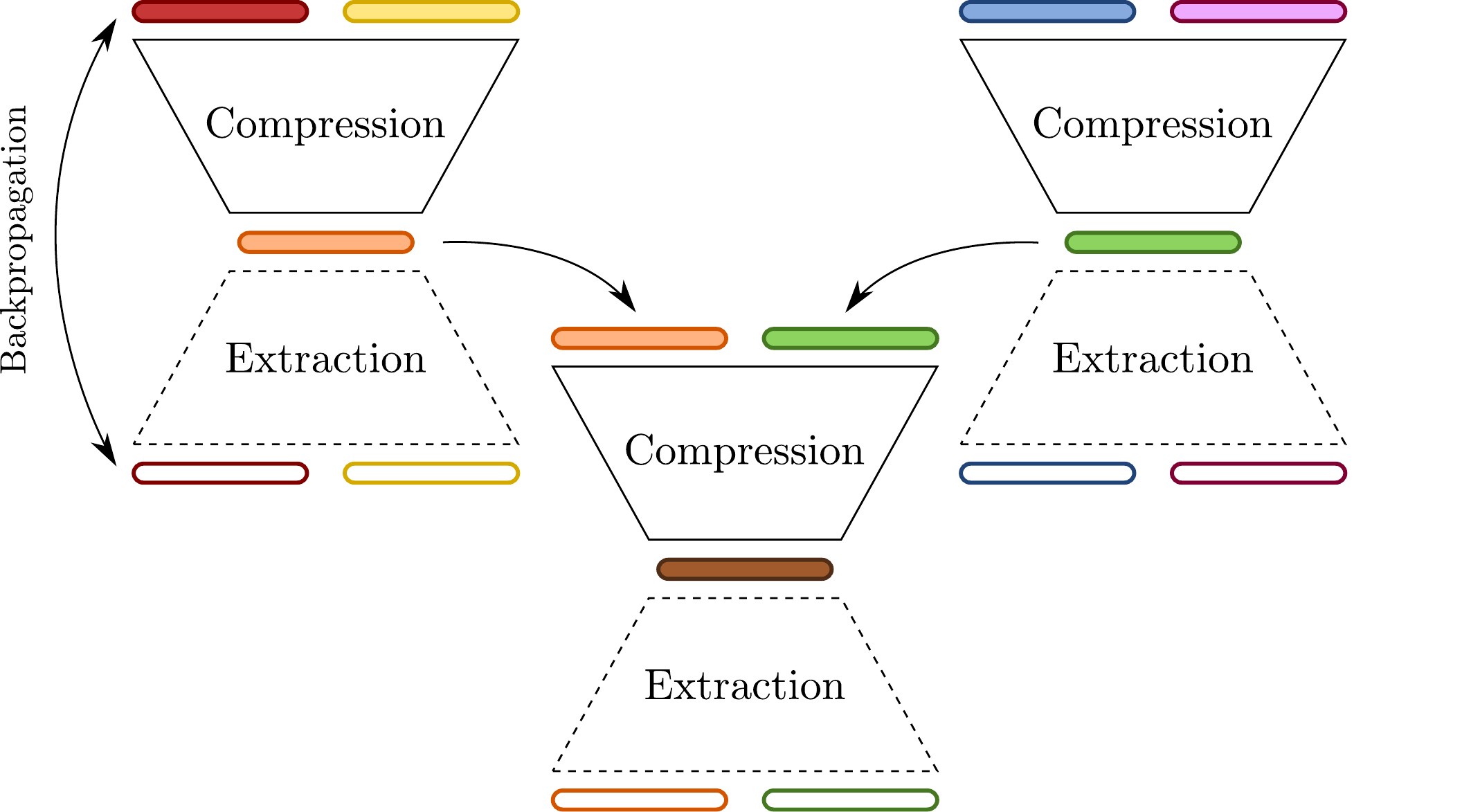}
\end{center}
\caption{Example of the training of the network with a sentence of length $l = 4$. The two first word vectors are forwarded through both compression and extraction part, and the mean squared error is backpropagated. The same is done with the third and fourth word vectors. Finally, the two outputs of the compression part obtained before are taken and forwarded through the auto-encoder, and the mean squared error is backpropagated. The output of the third compression is used as the encoded representation of the sentence.}
\label{fig:compression-training}
\end{figure}

We first tried to optimize our network locally: each time we compress two vectors $x,y$, we extract them back from their encoded representation, and want the extractions to be as close as possible to the original vectors. To do so, we use a mean squared error criterion, defined as follows:

$$L_w(x) = \frac{1}{d}\sum_{i=1}^d (x_i - f_i(x,w))^2$$

where $x$ is the concatenation of the two $d$-dimensional input vectors, and $f(x,w)$ the function that compresses and the extracts these two vectors, i.e. $f(x, w) = h^{-1}(h(x, w), w)$. This error is backpropagated through both the compression and extraction parts of the auto-encoder for each pair of vectors, thus $l - 1$ times per sentence, as illustrated in figure \ref{fig:compression-training}.

We chose to use this local approach because we felt that the global one (compressing a whole sentence to its encoded representation, then extracting it back and measuring the distance between the original words and the extracted ones) would be too hard to train. Indeed, a sentence of ten words (which is quite small) already implies to forward through nine compression networks and nine extraction networks. Backpropagating the error through this deep architecture would end up with very small gradients in the first layers, resulting in the first networks never to be trained. The local approach, however, is more like a greedy solution, that minimizes the error at each intermediate step of the compression, instead of doing it only once globally.
\\

The performance of our network is evaluated by looking at the mean squared distance between the input vectors of the auto-encoder and its output vectors, on a test set. While this measure indicates if the network is correctly training as we want, it does not give any information whether we could correctly extract the sentence or not. Thus, we used a second measure during the test phase: the proportion of correct extractions. This proportion is defined as the percentage of words for which the closest vector to the extracted word vector in the embedding space is the original word vector itself. Algorithm \ref{alg:correct-extractions} shows how it is computed.
\begin{algorithm}
\caption{Computation of the proportion of correct extractions.}
\begin{algorithmic}
\FOR{each sentence $s$}
\STATE{replace each word index $i$ of $s$ by the corresponding word vector $w_i$}
\STATE{compress $s$ to its encoded representation $z$}
\STATE{extract $s'$ back from $z$}
\FOR{each word vector $w'_i$ of $s'$}
\STATE{find the closest word vector $w_j$ to $w'_i$ in the embedding space}
\STATE{count how often $i = j$, i.e. the closet word is actually the original one}
\ENDFOR
\ENDFOR
\end{algorithmic}
\label{alg:correct-extractions}
\end{algorithm}

\subsubsection{Results}

When we obtained our first results, we observed that a decreasing mean squared error did not translate to better proportions of correct extractions. Besides, these proportions were good with short sentences (of size 2 or 3) but quickly decreased when considering longer sentences, suggesting that the greedy solution of optimizing the mean squared error at each step of the compression was not the right choice for having good extraction performances.

\subsection{Global optimization}

\begin{figure}
\begin{center}
\includegraphics[width=0.8\textwidth]{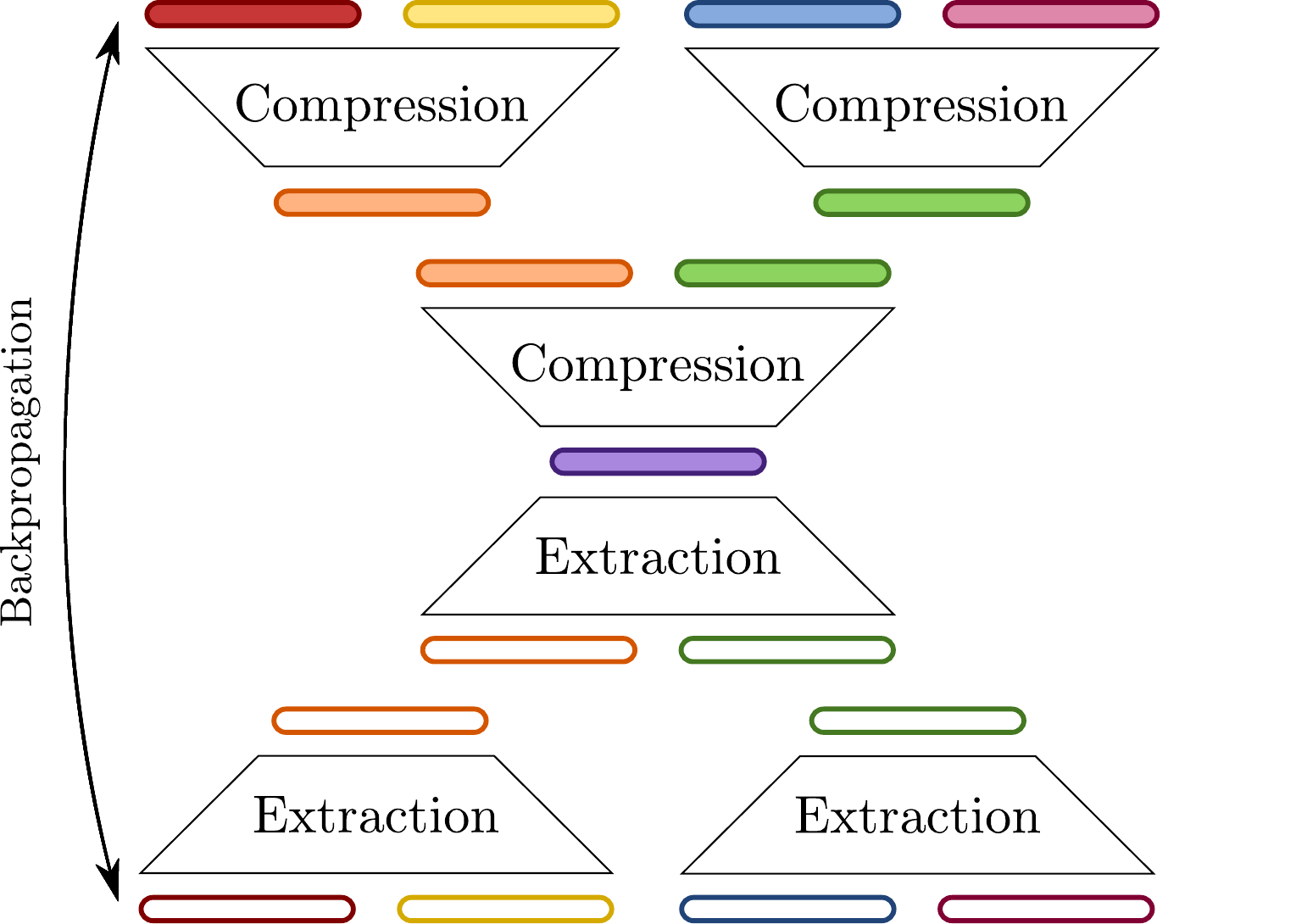}
\end{center}
\caption{Second version of the training scheme: contrary to the one pictured in figure \ref{fig:compression-training}, the error is only computed once between the original word vectors and the final extracted vectors, and then backpropagated directly through the whole network.}
\label{fig:compression-training2}
\end{figure}

Thus, we changed our training method to use an end-to-end technique: instead of considering each compression step separately and backpropagating the error locally, we performed the full compression and extraction process on the sentence, and evaluated the distance directly between the original word vectors and their corresponding extracted vectors. Then, we backpropagated this error through the whole extraction and compression networks. Figure \ref{fig:compression-training2} illustrates this new training scheme.

The main drawback of this end-to-end approach is that it limits the length of sentence that can be trained. Indeed, as explained above, the full processing of one sentence of length $l$ requires $l - 1$ compressions and the same number of extractions. This means that the error will be backpropagated $l - 1$ times through both the extraction and compression parts of our auto-encoder. The gradients decreasing at each step, the upper layers might never get significant gradients, and thus would never be learned.

\subsection{Results}

\begin{figure}
\begin{center}
\includegraphics[width=1\textwidth]{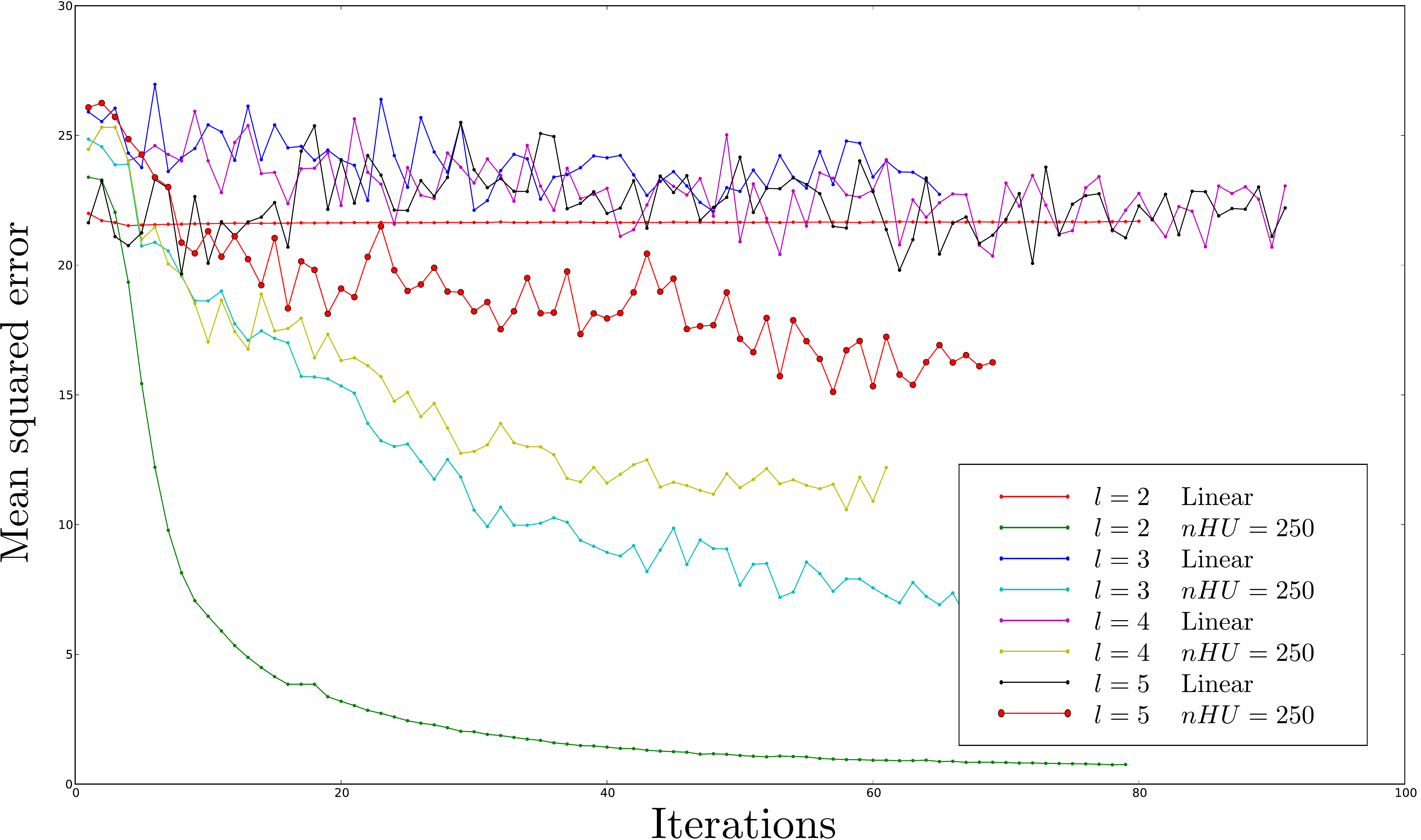}
\end{center}
\caption{Mean squared error in function of the number of iterations and with different configurations.}
\label{fig:compression-mse-results}
\end{figure}

\begin{figure}
\begin{center}
\includegraphics[width=1\textwidth]{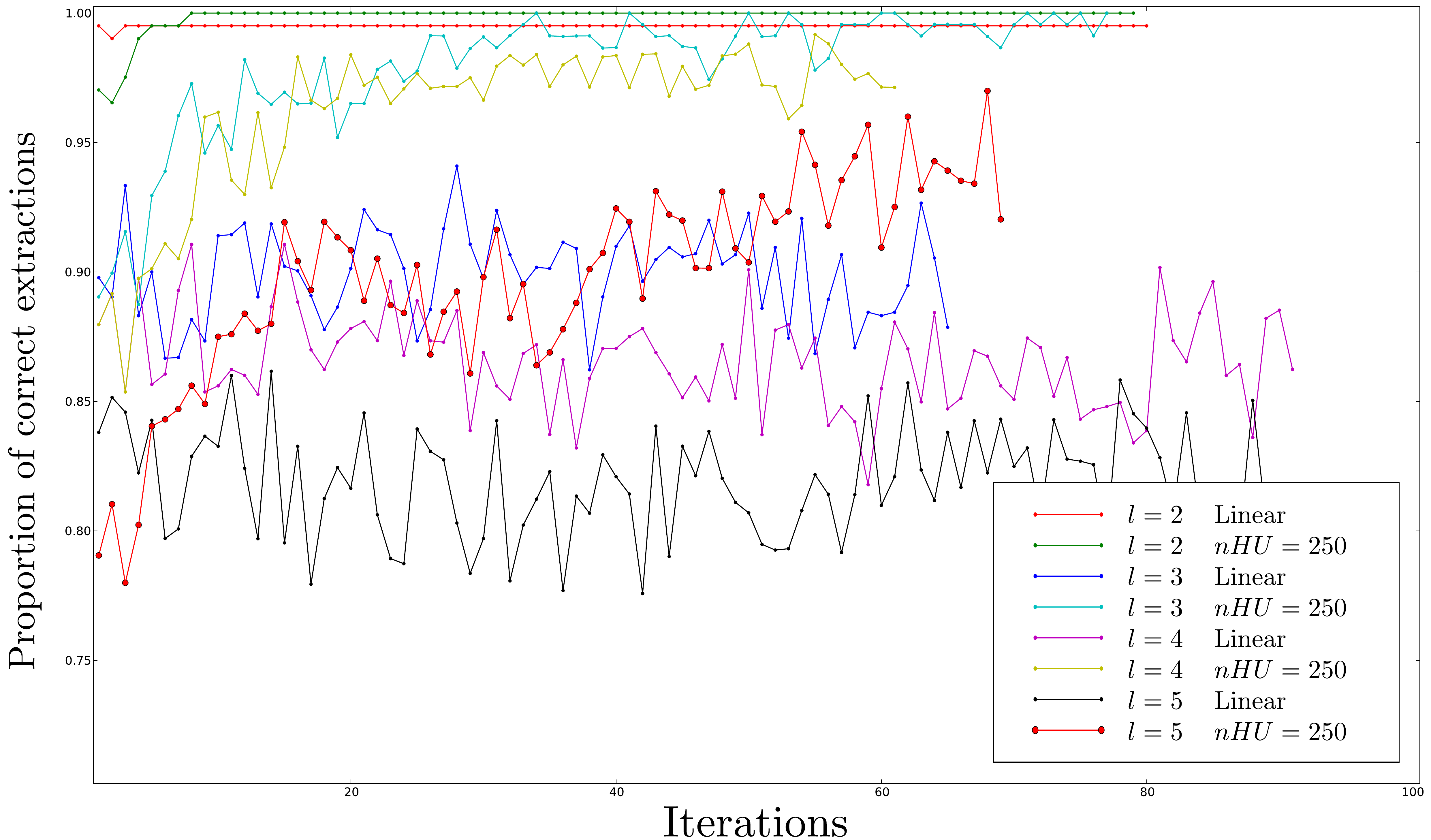}
\end{center}
\caption{Proportion of correct extractions of pairs of vectors using a mean-squared error criterion, in function of the number of iterations and with different configurations. }
\label{fig:compression-mse-uncompResults}
\end{figure}

Figure \ref{fig:compression-mse-results} shows the mean squared error between the original word vectors and the extracted ones, in function of the number of iterations, with both linear and non-linear networks, and for different sentences length $l$. Figure \ref{fig:compression-mse-uncompResults} shows the corresponding proportions of correct extractions. While our results are particularly noisy\footnote{Evaluating the proportion of correct extractions is pretty costly, because it implies computing the distance of each word of the dictionary to each extracted word vector. Thus, with 30'000 words in our dictionary, we had to use a testing set of limited size (100 paragraphs of approximatively 70 words in average), thus resulting in a big variance of our measure.}, we clearly see that decreasing the global mean squared error increases the proportion of correct extractions, especially for non-linear networks.

Moreover, while extracting nearly perfectly pairs of words, the linear auto-encoder quickly shows its limits when increasing the length of the sentence. The non-linear one is more difficult to train and thus takes longer to get good results, but is able to extract longer sentences with a much higher success rate. Again, by observing the slopes of the different curves, our graphs suggest that we could have obtained better results by training longer.
\\

However, we wanted to do some more testing before pursuing in this direction. Indeed, the pretty good extraction results seemed to imply that our network was actually doing what we wanted: learning the semantics. To verify this, we had a look at the embedding space and compared words to their closest neighbors, similarly to what we did in the previous chapter. If everything had worked well, it should have learned that, for example, "men" and "women" are similar concepts and thus have a similar representation in the embedding space. But the results were totally different: word embeddings made no sense, as if they had not moved from the randomly generated position they were assigned to during the initialization of our network.
\\

We suspect that instead of making the network learn the semantics, this criterion allowed it to get away with applying some tricks to its inputs (e.g. concatenating parts of the input vectors) and still obtain good results. We thus had to find another criterion that would compel the network to understand the meanings of the sentences, instead of simply applying some tricks on vectors.

\section{Second approach: Ranking}

Our second approach is based on a different way of training our network: instead of relying on the distance between the input and the output (this idea was inspired by auto-encoder networks), we introduce a notion of a score $score(z)$ attributed to the encoded representation $z$ of a sentence $s$. Then, we train the network such that the score of a sentence from the dataset is higher than the score of the same sentence with one word randomly replaced. This score can be seen as an indicator whether a given sentence is "correct" or not. By teaching the network how to differentiate the encoded representation of a correct sentence from the one of a wrong sentence, we should hopefully make it deduce an embedding of the words that makes sense.
\\

\subsection{Network architecture}

\begin{figure}
\begin{center}
\includegraphics[width=0.7\textwidth]{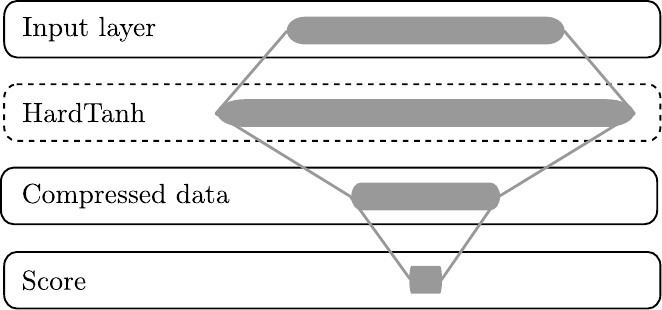}
\end{center}
\caption{Illustration of the last compression step with the ranking criterion: the compression layer (linear or not) is the same as before, and a linear layer with $d$ inputs and one output is simply appended next.}
\label{fig:ranking}
\end{figure}

The compression part is the same as the one we used so far, in both linear and non linear flavors, and the score function is simply implemented as a linear layer with $d$ inputs and one output, that is appended to the last compression step, as shown in figure \ref{fig:ranking}. Again, we train the network in a end-to-end manner, by forwarding both positive and negative examples through the network to obtain their encoded representations, computing their score, evaluating the error and finally backpropagating it through the whole network.

\subsection{Loss function}

Similarly to the first part of our work presented in the previous chapter, we used a margin ranking criterion for this learning task: 

$$L_w(x) = max(0, 1 - f_w(s_{pos}) + f_w(s_{neg}))$$

where $s_{pos}$ and $s_{neg}$ are respectively the correct and wrong sentence and $f_w(s)$ is the score of the encoded representation of a sentence $s$. 
\\

We started to group words differently from this point, using randomly generated trees in addition to left to right trees, to make sure that the network did not take advantage of any regularity in the way words were grouped and use some concatenation tricks to build the encoded representation. We also tried two different schemes to generated the "wrong sentences": either replacing always the last word of each sentence by a random word, or randomly choosing for each sentence which word to replace.
\\

To evaluate the performance of the network, we computed the rank of each sentence of the test set. This rank is defined similarly to the rank we used in the previous chapter: for a given sentence, we generate all possible negative samples by replacing one word by all the other words of the dictionary, and compute the score of each of these negative samples. Then, we sort them by decreasing score, and use the position of the correct sentence in the sorted list as its rank. Hence, a rank of one means that the correct sentence has a better score than all other negative samples. However, this cannot be achieved for all sentences, because some of the "negative" samples may actually be sentences that are correct, e.g. replacing the word "house" by "building" in a sentence will result in something grammatically correct in English, and should have a high score. Thus, we want the average rank of all sentences to be as high as possible, but cannot expect it to be one.

\subsection{Implementation}

We implemented these compression and extraction networks as modules in \texttt{Torch}. They allow to use any kind of tree for grouping words, and any kind of compression/extraction network. Moreover, we coded them following the architecture of a \texttt{Module} of the \texttt{nn} package (presented in chapter \ref{chap:torch}). This means that these compression/extraction trees can be included as modules in new or existing networks transparently, and can be trained the same way other modules are. Thus, it allows to easily build the kind of networks we used during our experiments, as illustrated in figure \ref{code:compresstree-sample}, without much effort for someone familiar with the \texttt{Torch} framework.

\begin{figure}
\begin{center}
\begin{lstlisting}
-- define the grouping tree for a sentence of length 6
tree = { { 1, { 2, 3 } }, { 4, { 5, 6 } } }

-- decompose tree in steps
steps = decompose(tree)

-- build compression part (function h)
mlpComp = nn.Linear(10, 5)

-- build the complete network as a sequence of three modules
mlp = nn.Sequential()

-- first module:
-- learns how to embed 100 words in a 5-dimensional space
mlp:add( nn.LookupTable(100, 5) ) 

-- second module:
-- creates a compression tree with the given compression
-- network and compression steps
mlp:add( nn.CompressTree( mlpComp, steps ) ) 

-- third module:
-- score layer
mlp:add( nn.Linear(5, 1) )

-- creating input:
-- sentence with words represented by their index in a dictionary
input = lab.new( 10, 4, 7, 29, 12, 84 )
                                 
-- get sentence score
score = mlp:forward( input )
\end{lstlisting}
\end{center}
\caption{Example of usage of our \texttt{CompressTree} module: we create a network composed of a lookup table (that maps word indices to vectors in 5 dimensions), a compression tree (that uses a linear compression network, \texttt{mlpComp}, and applies it in the order defined by \texttt{steps}) and a score layer (a simple linear layer that maps 5-dimensional vectors to a scalar). Then, we create a sentence and forward it through the network to get its score.}
\label{code:compresstree-sample}
\end{figure}

\subsection{Results}

\begin{figure}
\begin{center}
\includegraphics[width=1\textwidth]{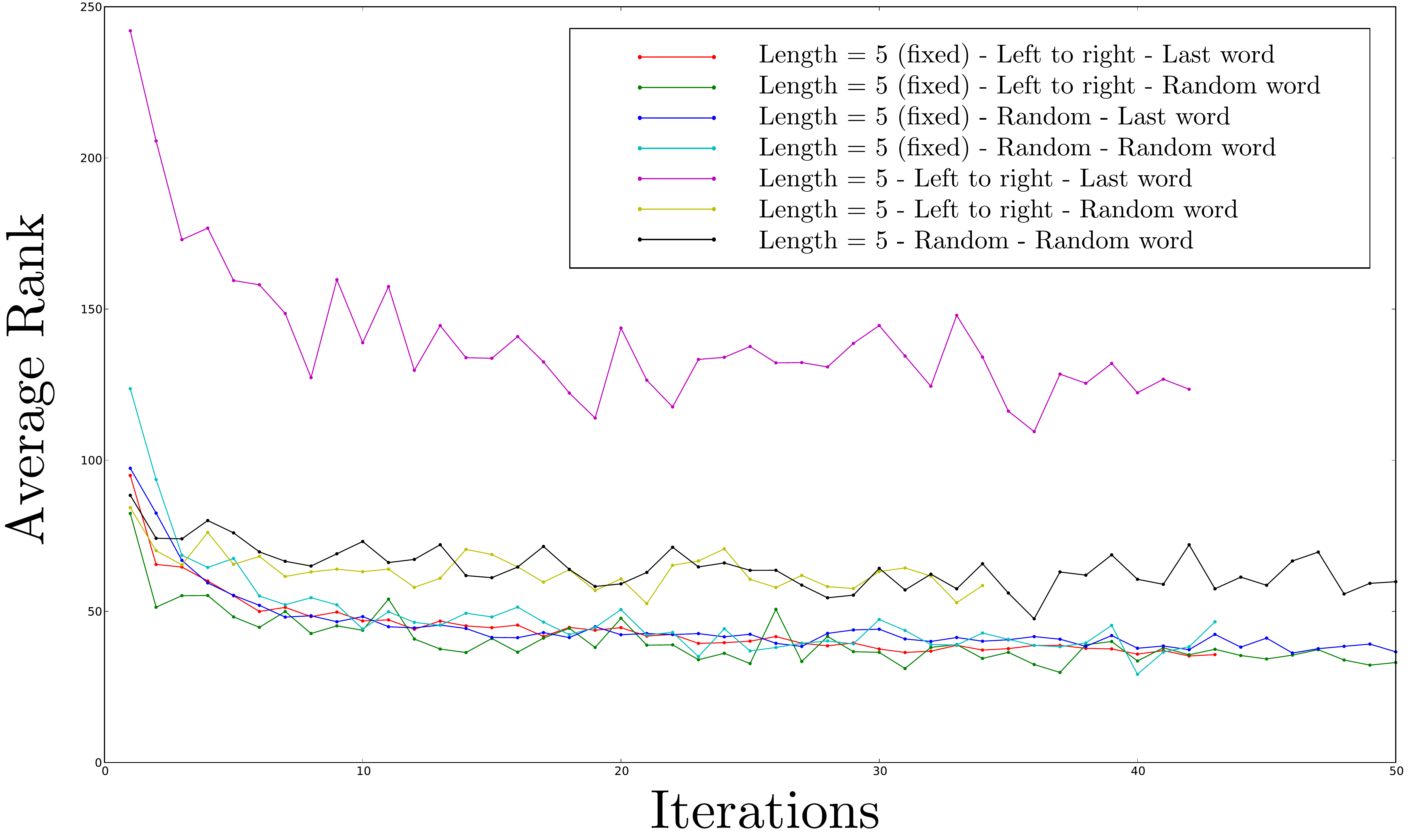}
\end{center}
\caption{Average rank of the sentences in the test set in function of the number of iterations, on a dictionary of 1000 words, with different ways of grouping words and generating negative samples. The compression part is the non-linear one with 250 hidden units, both trees and position of negative word are random, and the dimension of the embedding space 50.}
\label{fig:ranking-results}
\end{figure}

Figure \ref{fig:ranking-results} shows the results we obtained. We used an embedding dimension $d = 50$, a dictionary of 1000 words, a sentence length $l = 5$ (some configurations, marked as "fixed" on the graph, were using n-grams of length five only, whereas other used n-grams of random sizes between two and five) and the non-linear version of the compression network with two hundred and fifty hidden units, as this setup has shown some of the best results in the previous experiments while being small enough to train reasonably fast.
\\

The first observation that can be made by looking at these results is that left to right tree are no better than random trees. They both give the same kind of results in any configuration. However, this is a good thing, as it means that our network does not use any trick that depends on the regularity of the way words are combined. Moreover, training on n-grams of fixed length clearly gives better results than training on random lengths, or at least achieves them faster. The main reason for this is that using a fixed length always trains the network on the harder case, whereas random lengths will make it train for example on pairs of words one quarter of the time, which is really easier. Finally, results show that replacing always the last word to generate negative samples, or choosing which word to replace randomly for each sentence, does not influence the results. This is a major result, as previous attempts by Collobert et al., in the case of language models, were never able to learn similar tasks by replacing words randomly: they had to set a fixed position in order to make their network converge.
\\

Overall, the best performance we obtained was with a fixed length, random trees and choosing the position of the word to replace randomly to generate negative samples. In this setup, the best average rank was 25.45, with the worst rank being 1000. As explained before, we could not expect the average rank to be very close to one, as the negative samples generated may sometimes be actually correct sentences, and thus have a good score.
\\

\begin{figure}
\begin{center}
\includegraphics[width=1\textwidth]{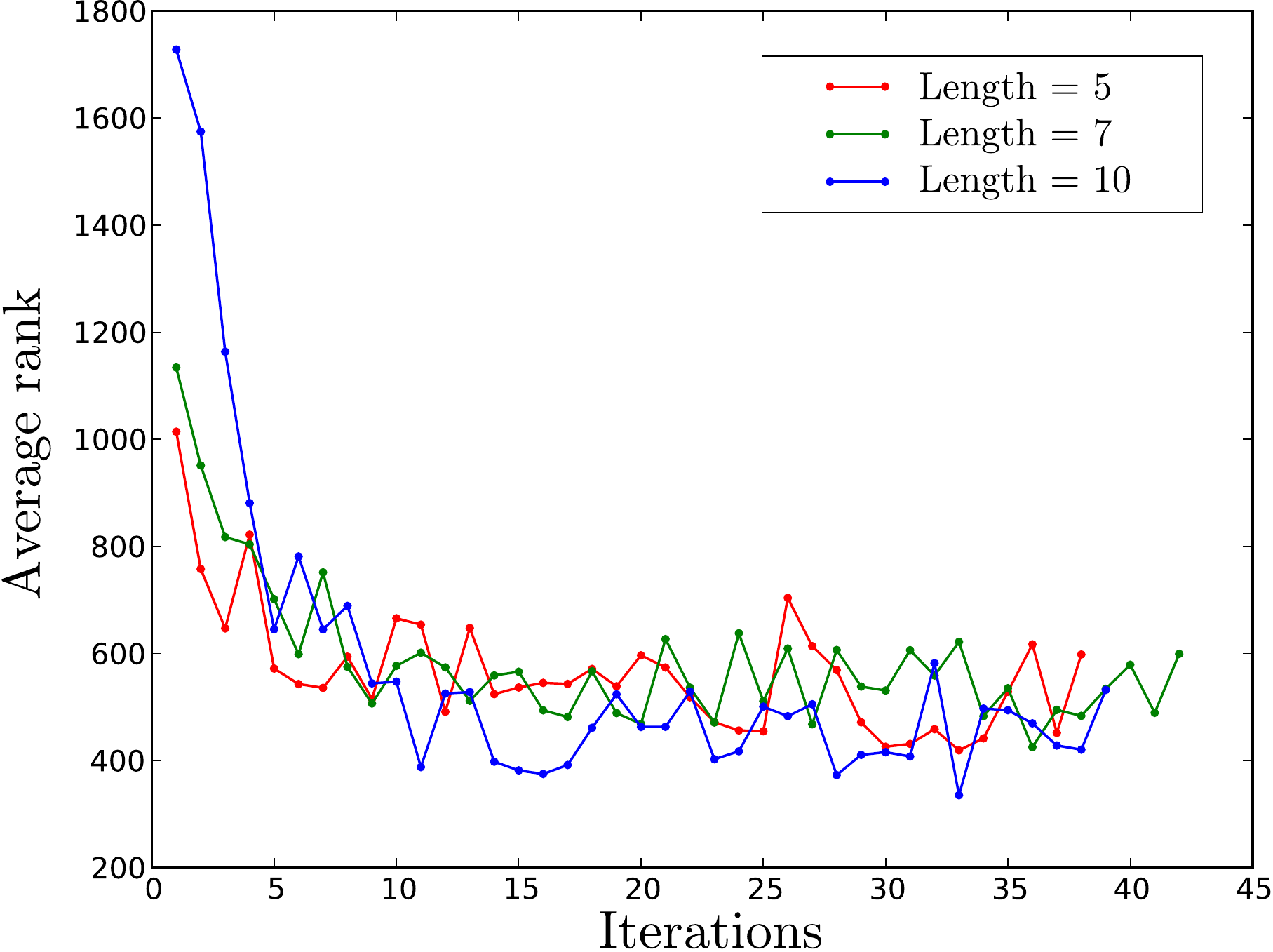}
\end{center}
\caption{Average rank of the sentences in the test set in function of the number of iterations, on a dictionary of 5000 words, with different n-grams lengths. The compression part is the non-linear one with 250 hidden units, both trees and position of negative word are random, and the dimension of the embedding space 50.}
\label{fig:ranking-results-lengths}
\end{figure}

Other n-grams lengths were also tried, to verify that the network was actually able to learn them and was not limited to small lengths. We again used an embedding dimension $d = 50$ and the non-linear compression network with $nHU = 250$ hidden units. This time, the dictionary size was 5'000 words, and we used random trees and a random position for changing one word in negative samples. The best average rank we obtained was a 335 with n-grams of length 10, which is quite encouraging. More detailed results are shown in figure \ref{fig:ranking-results-lengths}. While these results are comparatively worse than the one we obtained with a smaller dictionary, this is explained by the fact that this task is more difficult. Moreover, these experiments were run at the very end of our project, and the still slightly decreasing slope of the training errors suggests that there is still room for some improvement.

\subsection{A look at the resulting embedding}

Finally, we did some exploration in the embedding space, to see if the promising results obtained before actually meant that our network had built a sensible embedding for words. We first took some selected words, and looked for their closest neighbors in the embedding space (using the Euclidean metric), as presented in figure \ref{tab:ranking-results-words}. We see that words are semantically grouped, or at least are related. The embedding captures several features of the words, whether they are adjectives or nouns, singular or plural, etc. The same kind of results were obtained by Collobert et al. in \cite{word-embeddings}, but our architecture is more flexible: we both consider n-grams of fixed length while training, but where they only support replacing the middle word of the sentence, and use a kind of bag-of-word approach, we support any kind of grouping and replacing for our sentences.
\\

\begin{figure}
\begin{center}
\begin{tabular}{c|c|c|c|c|c}
two & year & red & city & women & france \\
\hline
three & week & blue & town & children & italy \\
four & day & black & village & artists & australia \\
six & season & green & church & men & germany \\
five & years & russian & state & characters & canada \\
eight & world & mountain & county & individuals & england \\
seven & post & gold & river & players & europe \\
ten & match & white & community & families & china \\
many & months & brown & area & people & paris \\
several & election & southern & university & groups & ireland \\
most & period & japanese & island & teams & japan \\
\end{tabular}
\end{center}
\caption{Selected words and their closest neighbors in the embedding space built by a network trained with n-grams of fixed length of 5, a dictionary of 1000 words, embedding space of dimension 50, random trees and random position of the word replace for generating random samples.}
\label{tab:ranking-results-words}
\end{figure}

Furthermore, while their language model can only assess relations between words, we can use our architecture to analyze relations between n-grams of any size. Basically, we only have to compute their encoded representation, and then are able to compare them directly, regardless if they represent a pair of words, a full sentence or even a whole document. The only caveat is that you have to compute all existing n-grams to find the closest one. This means that to find the closest pair of words to a given pair, if using a dictionary of $N$ words, we have to compute the score of all the $N^2$ different pairs and sort them to find the closest one. While this is feasible for n-grams of length two and small dictionaries, it is clearly too costly for longer sentences and real-world dictionaries. However, as a proof of concept, we selected a few pairs of words, and looked for the closest pairs in the embedding space using a limited dictionary. Figure \ref{tab:ranking-results-pairs} shows some selected pairs of words and their closest neighbors in the embedding space. Again, we see that they are have a close meaning, or at least are related.

\begin{figure}
\begin{center}
	\begin{tabular}{c|c|c|c}
last year & red house & the city & two men \\ 
\hline
first year & french house & the town & three men\\ 
same year &  rock house & the church & four men \\ 
first day & red court & the village &  two children \\ 
third year & german house & the state & two women \\ 
first season & black house & the country & three children
\end{tabular}
\end{center}
\caption{Selected pairs and their closest neighbors in the embedding space built by a network trained with n-grams of fixed length of 5, a dictionary of 1000 words, embedding space of dimension 50, random trees and random position of the word replace for generating random samples. The dictionary was limited to the 500 most frequent words when computing all the possible pairs.}
\label{tab:ranking-results-pairs}
\end{figure}

\section{Finding best tree}
\label{sec:compression-best-tree}

Finally, having a decent candidate for our compression function $h$, we tried to find the best way of grouping words. An important observation is that it is an iterative process: we start with a sentence of length $l$, select two consecutive words to group, apply $h$ and replace them by the resulting encoded representation. We repeat these operations $l - 1$ times, i.e. until we have only one vector left, which is the encoded representation of the sentence. This observation leads to an obvious greedy algorithm for grouping words. 

\subsection{Greedy algorithm}

\begin{figure}
\begin{center}
\includegraphics[width=0.6\textwidth]{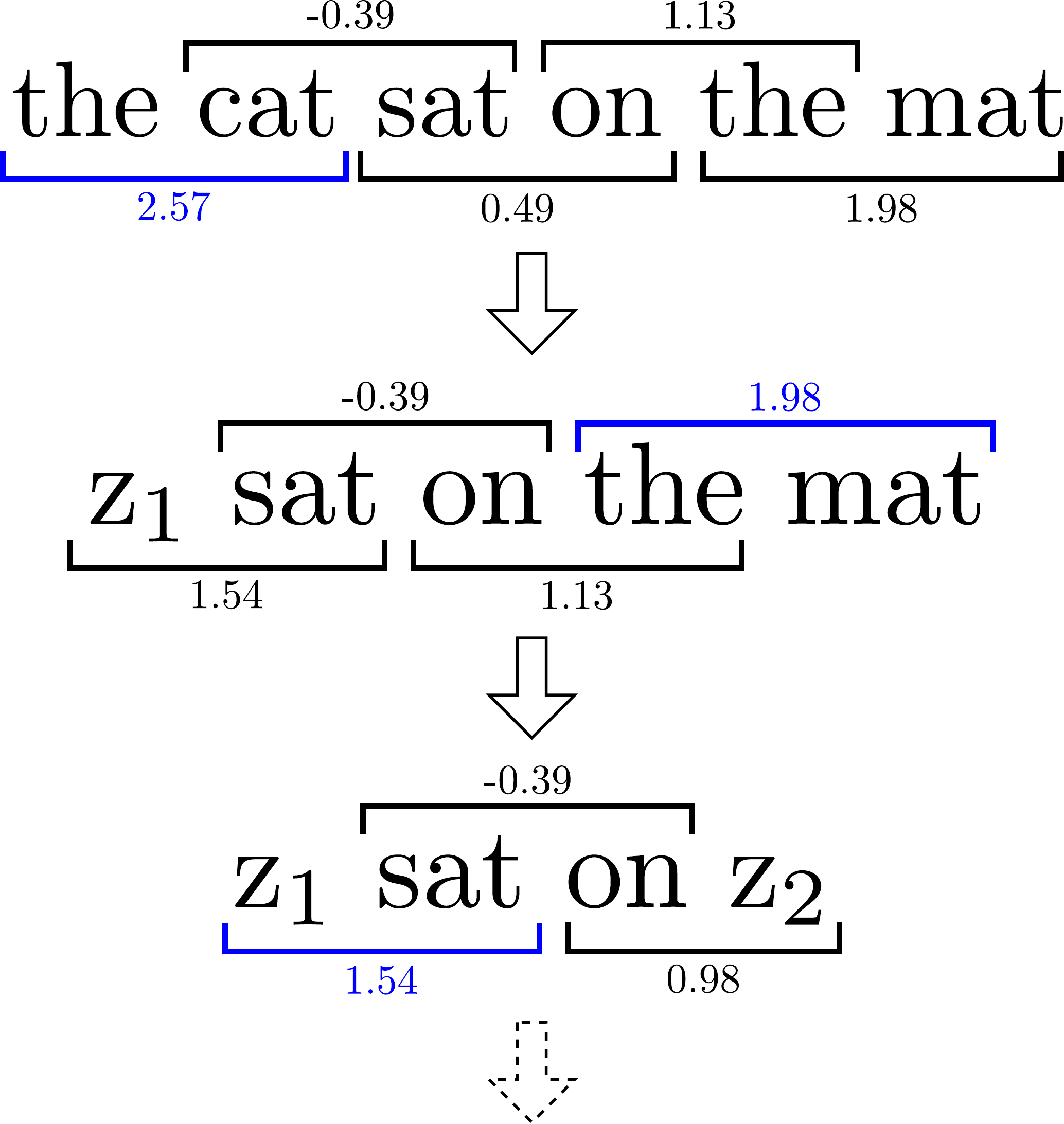}
\end{center}
\caption{Illustration of our greedy algorithm: when presented a sentence, compute the score of the compressed representations of all pairs of consecutive words, and select the best one (in blue). Replace the two words by their compressed representation, and repeat until the sentence has only one element left.}
\label{fig:greedy}
\end{figure}

For this algorithm, we make use of the notion of score introduced in the last section. This score can be seen as a mark attributed to an encoded representation that qualifies its quality: correct sentences should have a good encoded representation, and wrong sentences a bad one. Hence, we can make use of this score to build a simple algorithm that will greedily select at each step the best pair of words to group. This method is detailed in algorithm \ref{alg:best-tree} and illustrated in figure \ref{fig:greedy}.
\\

\begin{algorithm}
\caption{Greedy algorithm to determine how to group words when compressing a given sentence.}
\begin{algorithmic}
\WHILE{$s$ has more than one element}
\FOR{each pair of two consecutive elements of $s$}
\STATE{compute the score of the compressed representation of the pair}
\ENDFOR
\STATE{select the pair that has the best score}
\STATE{remove the two elements from $s$}
\STATE{insert their compressed representation}
\ENDWHILE
\end{algorithmic}
\label{alg:best-tree}
\end{algorithm}

\begin{figure}
\begin{center}
\includegraphics[width=0.7\textwidth]{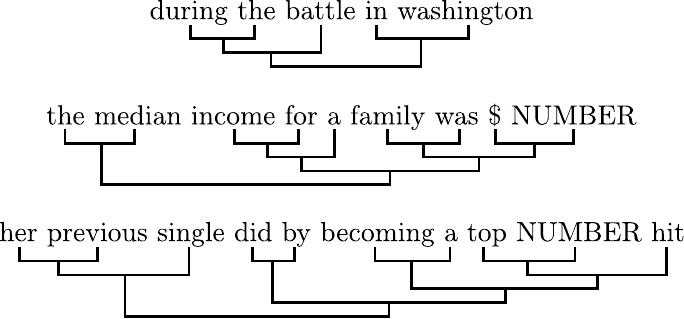}
\end{center}
\caption{Examples of the trees selected by our greedy algorithm on n-grams of different sizes.}
\label{fig:best-tree-results}
\end{figure}

We only had the opportunity to try our algorithm using a well-trained compression and ranking network in the very end of our project, and thus cannot yield many results nor really assess its performance.  We did however achieve an average rank of 16 when using the tree computed by our algorithm to compress sentences, and a dictionary of 1000 words, which is an encouraging results when compared to the best rank of 25 obtained with random trees. Figure \ref{fig:best-tree-results} shows some examples of the trees selected by our greedy algorithm when presented with different n-grams. Having no real mean of evaluating the correctness of such trees, we can only observe that most of the chosen groupings make sense.

\cleardoublepage
\chapter{Conclusion}
\label{chap:conclusion}

This chapter concludes our work. At the beginning of this document, we presented both Machine Translation and Sentence Compression fields, and explained the challenges awaiting us in both of them, and how we would try to take them up. We first succeeded in building a neural network that was able to learn how to translate from n-grams to n-grams, in any languages, provided that aligned sentences are available for training. However, it clearly showed its limitations when translating full sentences instead of simple n-grams. We will propose some extensions to try and circumvent this limitation in the next section, but translation remains, in our opinion, an open problem. Good translation tools will need to go a step further and start to actually understand the true meanings of what they are trying to translate.
\\

This is where the second part of our work comes into play. We built a flexible network that is able to embed any sentence into a $d$-dimensional space by iteratively applying a compression function to its elements two by two. It can use any kind of network for the compression phase, and combine the words in any order. While we were able to obtain good candidates for the compression function, we were not able to find the optimal way of combining words. Again, we will try to set some directions for improving this part in the next section. However, our architecture provides a good starting point for further experiments, being more powerful and flexible than traditional neural networks language models.

\section{Future work}

To conclude, we will now propose some ideas for continuing the work presented in this report. 

\subsection{Translating whole sentences}

The translation system we built in chapter \ref{chap:translation} is not really suited for translating whole sentences. However, if one still wanted to translate whole sentences, we can give here some hints about the direction to follow. As explained, our system can perform good n-grams to n-grams translations. Thus, when presentend with a sentence $s$ of length $l$ to translate, we could imagine taking all n-grams of size 1 to $l$, and computing the corresponding $k$ closest n-grams of the same size in the other language. Then, finding the best translation would only be a matter of selecting some of these n-grams, combining them, and maybe rearranging them (the order of the words may change from one language to the other), in order to be as close as possible to the original sentence in the embedding space.

\subsection{Improving the greedy algorithm}

In section \ref{sec:compression-best-tree}, we presented a simple version of a greedy algorithm that we used to find good ways of grouping words. While our approach has obtained a lower average rank than using random trees, it makes by construction choices that are locally optimal, and thus has few chances to end on a global optimum.
\\

We could instead design an algorithm that makes its decisions in a global way, based on the following observation: grouping words in a sentence can be seen as a sequence of decisions. Indeed, suppose we start with a stack that contains the first word of the sentence. We consider each word of the sentence successively, starting from the second one. At each step, we have two possible decisions: 
\begin{enumerate}
\item pop the stack, combine the current word with the one that was popped, and push the result of the combination on the stack
\item push the current word on the stack
\end{enumerate}

We can then attribute a cost to each of these choices, and run an algorithm similar to a Viterbi algorithm to find the globally optimal sequence of decisions. The costs themselves are still to be defined, but we feel that such an approach may achieve interesting results.
\cleardoublepage

\nocite{*}
\bibliographystyle{plain}
\bibliography{biblio} 
\addcontentsline{toc}{chapter}{Bibliography}

\end{document}